# A Trident Quaternion Framework for Inertial-based Navigation Part II: Error Models and Application to Initial Alignment

Wei Ouyang, Yuanxin Wu, *Senior Member*, *IEEE*

*Abstract*—**This work deals with error models for trident quaternion framework proposed in the companion paper (Part I) and further uses them to investigate the odometer-aided static/in-motion inertial navigation attitude alignment for land vehicles. By linearizing the trident quaternion kinematic equation, the left and right trident quaternion error models are obtained, which are found to be equivalent to those derived from profound group affine. The two error models are used to design their corresponding extended Kalman filters (EKF), namely, the left-quaternion EKF (LQEKF) and the right-quaternion EKF (RQEKF). Simulations and field tests are conducted to evaluate their actual performances. Owing to the high estimation consistency, the L/RQEKF converge much faster in the static alignment than the traditional error model-based EKF, even under arbitrary large heading initialization. For the in-motion alignment, the L/RQEKF possess much larger convergence region than the traditional EKF does, although they still require the aid of attitude initialization so as to avoid large initial attitude errors.**

## I. Introduction

THE coarse alignment is indispensable for a strapdown inertial navigation system (INS), in order to drive attitude errors to within several degrees [1], [2]. The main reason of this requirement resides in assuredly making valid the first-order linearization assumption of the extended Kalman filter (EKF) [3]. The initial alignment can be classified into the static alignment and the in-motion alignment, and the former particularly requires high-accuracy gyroscopes to sense the earth rotation for successful gyrocompassing [4]. In general, the performance of the subsequent fine alignment stage is largely determined by the rapidness and accuracy of the coarse alignment process.

The initial alignment methods can be generally classified into the vector observation-based ones and the Kalman filtering-based ones. As for the methods using finite vector observations, the initial attitude is directly determined with at least two non-collinear vectors, e.g., the three-axis attitude determination method (TRIAD) [1], [5], [6]. Moreover, the initial attitude can be obtained more robustly and accurately by the optimization-based alignment (OBA) approach proposed in [8], [9]. Both TRIAD and OBA are popular in static and in-motion alignment applications [10]-[15]. However, the optimal transition time from the coarse alignment to the fine alignment is non-trivial to be predetermined and the attitude covariance information is generally unavailable. In contrast, the Kalman filtering-based approach can estimate the attitude along with its error covariance. Nevertheless, the standard EKF necessitates a fine attitude initialization to obviate slow convergence or divergence since the linearization of the kinematic model is inaccurate under large attitude errors. To deal with this problem, researchers turned to use nonlinear angle-error models [16]-[18] or nonlinear filters such as the unscented Kalman filter [19], to account for large attitude errors. Unfortunately, these methods are mostly less effective than reported in practical implementation, regardless of the complicated computation.

Recently, an interesting route to deal with this problem is to circumscribe or even eliminate the EKF inconsistency caused by inappropriate linearization of the original nonlinear kinematic model. To achieve this goal, several effective methods have been proposed, of which the most influential approach is the invariant EKF (IEKF) in [20], [21]. Therein the system states evolve on Lie groups and new error definitions are accordingly obtained, making the resultant linearized error model estimate-invariant as long as the "group affine" condition is satisfied by the original system. The IEKF has been successfully used in improving the consistency of EKF-based strapdown INS [22][23], visual-inertial navigation [24]-[26], legged robot state estimation [27], to name but only a few. As claimed in [2], the IEKF can converge under large heading errors (e.g. 120 deg) in the static alignment, however, whether the IEKF could be directly used in the initial alignment from the very start is uncertain. In addition, the theoretical basis underneath the invariant theory is less friendly to novices and practitioners [28], [29]. Another way to improve the EKF consistency focuses on using the observability-constrained EKF (OC-EKF) proposed by Huang in [30], [31], which eliminates the unwanted "false observability" existing in the linearized models by null space linear projection. Nevertheless, the analytical solution and modification of the state transition matrix are nontrivial and sophisticated after incorporating the earth rotation in high-accuracy and long-distance situations.

This article tries to investigate the error models of the trident

The paper was supported in part by National Key R&D Program of China (2018YFB1305103) and National Natural Science Foundation of China (61673263).

W. Ouyang and Y. Wu are with Shanghai Key Laboratory of Navigation and Location-based Services, School of Electronic Information and Electrical Engineering, Shanghai Jiao Tong University, Shanghai 200240, China (email: ywoulife@sjtu.edu.cn, yuanx_wu@hotmail.com).

quaternion representation that is derived in the companion paper. Out of the same purpose of designing EKF algorithms with high consistency, the resultant error models and those under group affine [20], [21] are supposed to have enlightening connections. Starting with the newly proposed trident quaternion kinematic equation, the left- and right-error models are correspondingly derived. In contrast with [20], [21], the derivation process is much more direct and accessible, without requiring the profound knowledge of Lie group and differential manifold. Moreover, we will examine the derived models in the scenario of INS/odometer in-motion alignment, which is more challenging than the GPS-aided alignment, e.g. in [9] and [10].

The main contributions of this work reside in: 1) the trident quaternion based left/right-error models are explicitly derived; 2) the relation of the trident quaternion errors to the traditional state errors and the nonlinear state errors in [27], [35] is established; 3) simulations and field tests are performed to demonstrate the benefits of the trident quaternion-based left/right-error models in EKFs for the initial alignment under large initial heading uncertainties.

The remaining content is organized as follows. Section II derives the linearized trident quaternion kinematic equations for EKF, where the left/right-error models are presented. Subsequently, the relation of the trident quaternion errors to the traditional state errors and the nonlinear state errors is provided in Section III. The linearization of measurement models are derived in Section IV. Simulations are conducted in Section V for static and in-motion alignment, where the necessary backgrounds about INS/odometer integrated navigation system are provided. The results of land vehicle experiments are given in Section VI. Finally, Section VII concludes this article.

## II. LINEARIZED TRIDENT QUATERNION ERROR MODELS

In this section, the linearized trident quaternion kinematic models are derived. As shown in the companion paper, the vehicle's attitude, velocity and position can be compactly represented in one trident quaternion as

$$\breve{q}_{eb} = q_{eb} + \varepsilon_1 q'_{eb} + \varepsilon_2 q''_{eb}$$
$$= q_{eb} + \varepsilon_1 \frac{1}{2} \left( \mathbf{C}_i^e \dot{\mathbf{r}}^i \right) \circ q_{eb} + \varepsilon_2 \frac{1}{2} \left( \mathbf{C}_i^e \mathbf{r}^i \right) \circ q_{eb}, \quad (1)$$

where $q_{eb}$ is the attitude quaternion of the body frame relative to the earth frame. $q'_{eb}, q''_{eb}$ denote two dual parts of a trident quaternion. $\mathbf{C}_i^e$ is the attitude matrix from the inertial frame to the earth frame. $\dot{\mathbf{r}}^i$ and $\mathbf{r}^i$ are the inertial velocity and position, respectively. $\varepsilon_1, \varepsilon_2$ are two dual units, satisfying $\varepsilon_1^2 = \varepsilon_2^2 = \varepsilon_1 \varepsilon_2 = 0$ and $\varepsilon_1, \varepsilon_2 \neq 0$. The trident quaternion differential equation is given as

$$2\dot{\breve{q}}_{eb} = q_{eb} \circ \omega_{eb}^b + \varepsilon_1 \begin{bmatrix} \frac{1}{2} \left( \mathbf{C}_i^e \ddot{\mathbf{r}}^i \right) \circ q_{eb} \circ \omega_{ib}^b + q_{eb} \circ f^b \\ + g^e \circ q_{eb} - \frac{1}{2} \omega_{ie}^e \circ \left( \mathbf{C}_i^e \dot{\mathbf{r}}^i \right) \circ q_{eb} \end{bmatrix}$$
$$+ \varepsilon_2 \begin{bmatrix} \frac{1}{2} \left( \mathbf{C}_i^e \mathbf{r}^i \right) \circ q_{eb} \circ \omega_{ib}^b - \frac{1}{2} \omega_{ie}^e \circ \left( \mathbf{C}_i^e \mathbf{r}^i \right) \circ q_{eb} \\ + \left( \mathbf{C}_i^e \dot{\mathbf{r}}^i \right) \circ q_{eb} \end{bmatrix}. \quad (2)$$

which can be manipulated into the compact form

$$2\dot{\breve{q}}_{eb} = \breve{q}_{eb} \circ \breve{\omega}_{ib}^b - \breve{\omega}_{ie}^e \circ \breve{q}_{eb} \quad (3)$$

And, the solutions of two trident twists are assumed to take the form

$$\breve{\omega}_{ib}^b = \omega_{ib}^b + \varepsilon_1 f^b + \varepsilon_2 x_1$$
$$\breve{\omega}_{ie}^e = \omega_{ie}^e - \varepsilon_1 g^e + \varepsilon_2 x_2. \quad (4)$$

where $x_1 = [0, \mathbf{x}_1], x_2 = [0, \mathbf{x}_2], \mathbf{x}_1, \mathbf{x}_2 \in \mathbb{R}^3$ are usual vector quaternions.

Substitute Eq. (4) into Eq. (3) and compare it with Eq. (2), we have

$$(q_{eb} \circ x_1 - x_2 \circ q_{eb}) = \left( \mathbf{C}_i^e \dot{\mathbf{r}}^i \right) \circ q_{eb}, \quad (5)$$

Then the first-type special solutions of trident twists, which are used in the companion paper, are obviously obtained by letting $x_1$ be zero, i.e.,

$$\begin{cases} \breve{\omega}_{ib}^b = \omega_{ib}^b + \varepsilon_1 f^b \\ \breve{\omega}_{ie}^e = \omega_{ie}^e - \varepsilon_1 g^e - \varepsilon_2 \mathbf{C}_i^e \dot{\mathbf{r}}^i \end{cases}. \quad (6)$$

And, the second-type special solutions are obtained by letting $x_2$ be zero,

$$\begin{cases} \breve{\omega}_{ib}^b = \omega_{ib}^b + \varepsilon_1 f^b + \varepsilon_2 \mathbf{C}_i^e \dot{\mathbf{r}}^i \\ \breve{\omega}_{ie}^e = \omega_{ie}^e - \varepsilon_1 g^e \end{cases}. \quad (7)$$

In this work, only vectors are denoted in bold font with lowercase, such as $\mathbf{r}^i$. The 3-by-3 matrix is in uppercase bold font, such as $\mathbf{C}_i^e$. In order to be discriminated from attitude-related quaternions $q, q', q''$, vectors in italic font are used to denote vector quaternions, such as $\omega_{ib}^b = [0, \boldsymbol{\omega}_{ib}^b]$, $g^e = [0, \mathbf{g}^e]$, etc. In addition, when performing vector addition and multiplication with quaternions, 3-dimension vectors will be automatically transformed to vector quaternions by default; vector quaternions and vectors are interchangeable for the operation of vector product. For instance, in Eqs. (5) and (6) the vector $\mathbf{C}_i^e \dot{\mathbf{r}}^i$ has been treated as a vector quaternion $[0, \mathbf{C}_i^e \dot{\mathbf{r}}^i]$.

### A. Definitions of Left and Right Attitude errors

Attitude error is a relative concept and can be alternatively attributed to either of the two related frames. Hereby, the true attitude matrix is denoted in $\mathbf{C}_b^e$, and it can be related with its estimation as $\mathbf{C}_b^e = \mathbf{C}_{\hat{b}}^e \mathbf{C}_b^{\hat{b}}$ or alternatively $\mathbf{C}_b^e = \mathbf{C}_{\hat{e}}^e \mathbf{C}_b^{\hat{e}}$, where '^' denotes the erroneous or estimated frame. The former relation totally owes the attitude error between the Earth/body frames to the erroneous body frame, while the latter relation owes it to the erroneous Earth frame. Under small attitude error, the error matrices can be approximated as $\mathbf{C}_b^{\hat{b}} \approx \mathbf{I} + \Delta\boldsymbol{\sigma}_l \times$ and $\mathbf{C}_{\hat{e}}^e \approx \mathbf{I} + \Delta\boldsymbol{\sigma}_r \times$, respectively. The symbol "$\times$" denotes the antisymmetric operation. Note that the in the subsequent derivations we indiscriminately use $\hat{\mathbf{C}}_b^e$ to denote the estimated attitude matrix $\mathbf{C}_b^{\hat{e}}$ or $\mathbf{C}_{\hat{b}}^e$, if not particularly clarified.

When using the quaternion $q_{eb}$ to express the attitude, we define the left quaternion error and the right quaternion error as

$$\Delta q_l = q_{b\hat{b}} = q_{eb}^* \circ q_{e\hat{b}} \approx [1, \tfrac{1}{2}\Delta\boldsymbol{\sigma}_l]$$
$$\Delta q_r = q_{\hat{e}e} = q_{e\hat{b}} q_{eb}^* \circ \approx [1, \tfrac{1}{2}\Delta\boldsymbol{\sigma}_r]. \tag{8}$$

in which, we can see that the first-order Euler angle error are twice of the vector part of the quaternion error. In the following derivations, the subscripts of quaternions are neglected for brevity. Additionally, the unit quaternion is defined as $1 = [1, \mathbf{0}]$ and the unit trident quaternion is defined as $\bar{1} = [1, \mathbf{0}] + \varepsilon_1[0, \mathbf{0}] + \varepsilon_2[0, \mathbf{0}]$ in the sequel.

It should be underlined that the above left/right error concept is not utterly new to the navigation community. As done in [36], [37], for example, the attitude error was alternatively treated in the reference frame or in the body frame while deriving the error model of the strapdown INS.

### B. Left Trident Quaternion Error Model

The left trident quaternion error to the first order is defined as

$$\Delta\bar{q}_l = \hat{\bar{q}}^* \circ \bar{q} \approx 1 + \tfrac{1}{2}\Delta\bar{\sigma}_l, \tag{9}$$

Note here '1' actually denotes the unit trident quaternion, and the error term is also expressed in the vector trident quaternion as

$$\Delta\bar{\sigma}_l = \Delta\boldsymbol{\sigma}_l + \varepsilon_1\Delta\boldsymbol{\sigma}_l' + \varepsilon_2\Delta\boldsymbol{\sigma}_l'', \tag{10}$$

where the vector quaternions are specifically given as $\Delta\boldsymbol{\sigma}_l = [0, \Delta\boldsymbol{\sigma}_l]$, $\Delta\boldsymbol{\sigma}_l' = [0, \Delta\boldsymbol{\sigma}_l']$ and $\Delta\boldsymbol{\sigma}_l'' = [0, \Delta\boldsymbol{\sigma}_l'']$.

Accordingly, the derivative of the left trident quaternion error is computed as

$$\begin{aligned}
\Delta\dot{\bar{\sigma}}_l &= 2d\,\hat{\bar{q}}^*/dt \circ \bar{q} + 2\hat{\bar{q}}^* \circ \dot{\bar{q}} \\
&= -\hat{\bar{q}}^* \circ \left(\dot{\hat{\bar{q}}} \circ \hat{\bar{\omega}}_{ib}^b - \hat{\bar{\omega}}_{ie}^e \circ \hat{\bar{q}}\right) \circ \hat{\bar{q}}^* \circ \bar{q} \\
&\quad + \hat{\bar{q}}^* \circ \left(\bar{q} \circ \bar{\omega}_{ib}^b - \bar{\omega}_{ie}^e \circ \bar{q}\right) \\
&= -\hat{\bar{\omega}}_{ib}^b \circ \hat{\bar{q}}^* \circ \bar{q} + \hat{\bar{q}}^* \circ \bar{q} \circ \bar{q}^* \circ \hat{\bar{\omega}}_{ie}^e \circ \bar{q} \\
&\quad + \hat{\bar{q}}^* \circ \bar{q} \circ \left(\bar{\omega}_{ib}^b - \bar{q}^* \circ \bar{\omega}_{ie}^e \circ \bar{q}\right).
\end{aligned} \tag{11}$$

where the formula $d\bar{q}^*/dt = -\bar{q}^* \circ d\bar{q}/dt \circ \bar{q}^*$ is used.

Using (4), it can be obtained that

$$\begin{aligned}
\Delta\dot{\bar{\sigma}}_l &= -\boldsymbol{\omega}_{ib}^b \times \Delta\boldsymbol{\sigma}_l - \delta\boldsymbol{\omega}_{ib}^b - \varepsilon_1 \begin{pmatrix} \mathbf{C}_e^b \delta\mathbf{g}^e + f^b \times \Delta\boldsymbol{\sigma}_l \\ + \delta f^b + \boldsymbol{\omega}_{ib}^b \times \Delta\boldsymbol{\sigma}_l' \end{pmatrix} \\
&\quad - \varepsilon_2\left(\mathbf{x}_1 - \mathbf{C}_e^b\delta\mathbf{x}_2 + x_1 \times \Delta\boldsymbol{\sigma}_l + \boldsymbol{\omega}_{ib}^b \times \Delta\boldsymbol{\sigma}_l''\right).
\end{aligned} \tag{12}$$

of which the detailed derivations are provided in Appendix A.

It can be seen that only the second dual part of Eq. (12) contains the unknown input quaternions $x_1$ and $\delta\mathbf{x}_2 \in \mathbb{R}^3$. In order to linearize the second dual part w.r.t. the trident quaternion errors and make the resultant coefficients time-

invariant, we hope that the following condition in terms of the second dual part in Eq. (12) can be fulfilled

$$\mathbf{x}_1 - \mathbf{C}_e^b\delta\mathbf{x}_2 + \mathbf{x}_1 \times \Delta\boldsymbol{\sigma}_l = \mathbf{k}_1\Delta\boldsymbol{\sigma}_l + \mathbf{k}_2\Delta\boldsymbol{\sigma}_l' + \mathbf{k}_3\Delta\boldsymbol{\sigma}_l''. \tag{13}$$

in which, $\mathbf{k}_1, \mathbf{k}_2, \mathbf{k}_3 \in \mathbb{R}^{3\times3}$ are time-invariant real matrices.

In addition, for the left trident quaternion error, its dual parts are further denoted as the error-state vectors (see Appendix B for details)

$$\begin{aligned}
\Delta\boldsymbol{\sigma}_l' &= -\mathbf{C}_e^b\delta\left(\mathbf{C}_i^e\dot{\mathbf{r}}^i\right) \\
\Delta\boldsymbol{\sigma}_l'' &= -\mathbf{C}_e^b\delta\left(\mathbf{C}_i^e\mathbf{r}^i\right).
\end{aligned} \tag{14}$$

Substitute Eq. (14) into (13), we have

$$\mathbf{x}_1 - \mathbf{C}_e^b\delta\mathbf{x}_2 + \mathbf{x}_1 \times \Delta\boldsymbol{\sigma}_l = \mathbf{k}_1\Delta\boldsymbol{\sigma}_l - \mathbf{k}_2\mathbf{C}_e^b\delta\left(\mathbf{C}_i^e\dot{\mathbf{r}}^i\right) - \mathbf{k}_3\mathbf{C}_e^b\delta\left(\mathbf{C}_i^e\mathbf{r}^i\right). \tag{15}$$

Instead of investigating all possible solutions, two groups of apparent special solutions are directly examined. Substitute Eqs. (6) and (7) into the above condition, and it can be readily found that the first-type solutions are valid with $\mathbf{k}_1 = \mathbf{0}, \mathbf{k}_2 = \mathbf{I}, \mathbf{k}_3 = \mathbf{0}$. And, the second dual part in Eq. (12) can be finally denoted as

$$\begin{aligned}
\Delta\dot{\boldsymbol{\sigma}}_l'' &= -\mathbf{C}_e^b\delta\left(\mathbf{C}_i^e\dot{\mathbf{r}}^i\right) - \boldsymbol{\omega}_{ib}^b \times \Delta\boldsymbol{\sigma}_l'' \\
&= \Delta\boldsymbol{\sigma}_l' - \boldsymbol{\omega}_{ib}^b \times \Delta\boldsymbol{\sigma}_l''.
\end{aligned} \tag{16}$$

Note that the pivotal motivation to find the invariant linearization is to improve the consistency of the resultant EKF, which is not affected by inaccurate estimations.

In order to linearize the remaining two parts, the body frame angular velocity measurements are explicitly defined as $\tilde{\boldsymbol{\omega}}_{ib}^b = \boldsymbol{\omega}_{ib}^b + \mathbf{b}_g + \mathbf{n}_g$, and the measured specific forces are $\tilde{\mathbf{f}}^b = \mathbf{f}^b + \mathbf{b}_a + \mathbf{n}_a$. Accordingly, their error terms are taken as the corrected estimates minus the truth

$$\delta\boldsymbol{\omega}_{ib}^b = \left(\tilde{\boldsymbol{\omega}}_{ib}^b - \hat{\mathbf{b}}_g\right) - \boldsymbol{\omega}_{ib}^b = -\delta\mathbf{b}_g + \mathbf{n}_g, \tag{17}$$

$$\delta\mathbf{f}^b = \left(\tilde{\mathbf{f}}^b - \hat{\mathbf{b}}_a\right) - \mathbf{f}^b = -\delta\mathbf{b}_a + \mathbf{n}_a. \tag{18}$$

According to Appendix B, we have $\delta\mathbf{r}^e = -\mathbf{C}_b^e\Delta\boldsymbol{\sigma}_l''$, and thus, the gravitational force error in Eq. (12) can be computed as

$$\delta\mathbf{g}^e = \hat{\mathbf{g}}^e - \mathbf{g}^e = \frac{\partial\mathbf{g}^e}{\partial\mathbf{r}^e}\delta\mathbf{r}^e = -\frac{\partial\mathbf{g}^e}{\partial\mathbf{r}^e}\mathbf{C}_b^e\Delta\boldsymbol{\sigma}_l'', \tag{19}$$

in which, the term $\partial\mathbf{g}^e/\partial\mathbf{r}^e$ is approximated as (see [4], ch. 14)

$$\frac{\partial\mathbf{g}^e}{\partial\mathbf{r}^e} \approx -\frac{2\mathbf{g}^e}{\hat{\mathbf{r}}_{eS}^e} \frac{\hat{\mathbf{r}}^{eT}}{\left\|\hat{\mathbf{r}}^e\right\|}. \tag{20}$$

where the formulae for the geocentric radius $\hat{\mathbf{r}}_{eS}^e$ and the gravitational force $\mathbf{g}^e$ can also be found in [4] (ch. 2).

Substitute Eqs. (16)-(19) into (12) and rewrite the differential equation of trident quaternion error in matrix form, the linearized left-error model is finally united in

$$\delta\dot{\mathbf{x}}_l = \mathbf{F}_l\delta\mathbf{x}_l + \mathbf{G}_l\mathbf{w}_k. \tag{21}$$

In Eq. (21), the error states, Jacobian matrices and noises are given as

$$\delta\mathbf{x}_l = \begin{bmatrix} \Delta\boldsymbol{\sigma}_l & \Delta\boldsymbol{\sigma}_l' & \Delta\boldsymbol{\sigma}_l'' & \delta\mathbf{b}_g & \delta\mathbf{b}_a \end{bmatrix}^T \tag{22}$$

$$\mathbf{w}_k = \begin{bmatrix} \mathbf{w}_{grw} & \mathbf{w}_{arw} & \mathbf{n}_g & \mathbf{n}_a \end{bmatrix}^T \tag{23}$$

where $\mathbf{w}_{grw}$ and $\mathbf{w}_{arw}$ are the random walks of gyroscopes and accelerometers, respectively; $\mathbf{n}_g$ and $\mathbf{n}_a$ denote the noises of gyroscope bias and accelerometer bias.

$$\mathbf{F}_l = \begin{bmatrix} -\boldsymbol{\omega}_{ib}^b \times & \mathbf{0} & \mathbf{0} & \mathbf{I} & \mathbf{0} \\ -\mathbf{f}^b \times & -\boldsymbol{\omega}_{ib}^b \times & \mathbf{C}_e^b \dfrac{\partial \mathbf{g}^e}{\partial \mathbf{r}^e} \mathbf{C}_b^e & \mathbf{0} & \mathbf{I} \\ \mathbf{0} & \mathbf{I} & -\left(\boldsymbol{\omega}_{ib}^b \times\right) & \mathbf{0} & \mathbf{0} \\ \mathbf{0} & \mathbf{0} & \mathbf{0} & \mathbf{0} & \mathbf{0} \\ \mathbf{0} & \mathbf{0} & \mathbf{0} & \mathbf{0} & \mathbf{0} \end{bmatrix} \quad (24)$$

$$\mathbf{G}_l = \begin{bmatrix} -\mathbf{I} & \mathbf{0} & \mathbf{0} & \mathbf{0} \\ \mathbf{0} & -\mathbf{I} & \mathbf{0} & \mathbf{0} \\ \mathbf{0} & \mathbf{0} & \mathbf{0} & \mathbf{0} \\ \mathbf{0} & \mathbf{0} & \mathbf{I} & \mathbf{0} \\ \mathbf{0} & \mathbf{0} & \mathbf{0} & \mathbf{I} \end{bmatrix} \quad (25)$$

*Remark 1:* In the Jacobian matrix $\mathbf{F}_l$, the term related to the gravitational gradient is in the order of $10^{-8}$ in general, therefore, the left trident quaternion error model can be regarded as free from inaccurate estimations over the linearization interval.

### C. Right Trident Quaternion Error Model

The right trident quaternion error is defined as

$$\Delta \hat{\bar{q}}_r = \bar{q} \circ \hat{\bar{q}}^* \approx 1 + \frac{1}{2} \Delta \bar{\sigma}_r, \quad (26)$$

And, the right trident quaternion error is further denoted as

$$\Delta \bar{\sigma}_r = \Delta \sigma_r + \varepsilon_1 \Delta \sigma_r' + \varepsilon_2 \Delta \sigma_r'', \quad (27)$$

in which, the vector quaternions are defined as $\Delta \sigma_r = [0, \Delta \boldsymbol{\sigma}_r]$, $\Delta \sigma_r' = [0, \Delta \boldsymbol{\sigma}_r']$ and $\Delta \sigma_r'' = [0, \Delta \boldsymbol{\sigma}_r'']$.

The differential equation of the right trident quaternion error is accordingly derived as

$$\begin{aligned} \Delta \dot{\bar{\sigma}}_r &\approx 2\Delta \dot{\hat{q}}_r = 2\dot{\bar{q}} \circ \hat{\bar{q}}^* + 2\bar{q} \circ d\hat{\bar{q}}^* \big/ dt \\ &= \left( \bar{q} \circ \bar{\omega}_{ib}^b - \bar{\omega}_{ie}^e \circ \bar{q} \right) \circ \hat{\bar{q}}^* - \bar{q} \circ \left[ \hat{\bar{q}}^* \circ \left( \hat{\bar{q}} \circ \hat{\bar{\omega}}_{ib}^b - \hat{\bar{\omega}}_{ie}^e \circ \hat{\bar{q}} \right) \circ \hat{\bar{q}}^* \right] \\ &= \left( \bar{q} \circ \bar{\omega}_{ib}^b \circ \bar{q}^* - \bar{\omega}_{ie}^e \right) \circ \Delta \bar{q}_r - \bar{q} \circ \hat{\bar{\omega}}_{ib}^b \circ \bar{q}^* \circ \Delta \bar{q}_r + \Delta \bar{q}_r \circ \hat{\bar{\omega}}_{ie}^e. \end{aligned} \quad (28)$$

It can be further obtained that

$$\begin{aligned} \Delta \dot{\bar{\sigma}}_r &= \Delta \sigma_r \times \omega_{ie}^e - \mathbf{C}_b^e \delta \omega_{ib}^b \\ &\quad - \varepsilon_1 \begin{pmatrix} \omega_{ie}^e \times \Delta \sigma_r' - g^e \times \Delta \sigma_r + \delta g^e \\ + \left( \mathbf{C}_i^e \hat{\mathbf{r}}^i \right) \times \mathbf{C}_b^e \delta \omega_{ib}^b + \mathbf{C}_b^e \delta \mathbf{f}^b \end{pmatrix} \\ &\quad - \varepsilon_2 \begin{pmatrix} \mathbf{r}^e \times \mathbf{C}_b^e \delta \omega_{ib}^b + \delta x_2 - x_2 \times \Delta \sigma_r \\ + \mathbf{C}_b^e \delta \mathbf{x}_1 + \omega_{ie}^e \times \Delta \sigma_r'' \end{pmatrix}. \end{aligned} \quad (29)$$

for which $\delta x_2 = [0, \delta \mathbf{x}_2]$ and detailed derivations are provided in Appendix C.

Let us focus on the linearization of the second dual part. Similarly, it is expected to satisfy the following condition so as to get time-invariant coefficients w.r.t. the trident quaternion errors

$$\delta \mathbf{x}_2 - \mathbf{x}_2 \times \Delta \boldsymbol{\sigma}_r + \mathbf{C}_b^e \delta \mathbf{x}_1 = \mathbf{k}_1 \Delta \boldsymbol{\sigma}_r + \mathbf{k}_2 \Delta \boldsymbol{\sigma}_r' + \mathbf{k}_3 \Delta \boldsymbol{\sigma}_r''. \quad (30)$$

in which, $\mathbf{k}_1, \mathbf{k}_2, \mathbf{k}_3 \in \mathbb{R}^{3 \times 3}$ are time-invariant matrices.

Substitute the following relationships in Eq. (31) (see the derivation of Eq. (100)) into Eq. (30),

$$\begin{aligned} \Delta \boldsymbol{\sigma}_r' &= \left( \mathbf{C}_i^e \hat{\mathbf{r}}^i \right) \times \Delta \boldsymbol{\sigma}_r - \delta \left( \mathbf{C}_i^e \hat{\mathbf{r}}^i \right) \\ \Delta \boldsymbol{\sigma}_r'' &= \left( \mathbf{C}_i^e \hat{\mathbf{r}}^i \right) \times \Delta \boldsymbol{\sigma}_r - \delta \left( \mathbf{C}_i^e \hat{\mathbf{r}}^i \right) \\ &= \mathbf{r}^e \times \Delta \boldsymbol{\sigma}_r - \delta \mathbf{r}^e. \end{aligned} \quad (31)$$

and examine the two groups of special solutions in Eqs. (6) and (7), respectively. It indicates that the first-type solutions are still valid and $\mathbf{k}_1 = \mathbf{0}, \mathbf{k}_2 = -\mathbf{I}, \mathbf{k}_3 = \mathbf{0}$. As a result, the second dual part in Eq. (29) is computed as

$$\Delta \dot{\boldsymbol{\sigma}}_r'' = \mathbf{r}^e \times \mathbf{C}_b^e \delta \boldsymbol{\omega}_{ib}^b + \Delta \boldsymbol{\sigma}_r' + \boldsymbol{\omega}_{ie}^e \times \Delta \boldsymbol{\sigma}_r''. \quad (32)$$

Next, we further consider the term related to the gravitational error.

$$\begin{aligned} \delta \mathbf{g}^e &= \frac{\partial \mathbf{g}^e}{\partial \mathbf{r}^e} \delta \mathbf{r}^e \\ &= \frac{\partial \mathbf{g}^e}{\partial \mathbf{r}^e} \left( \mathbf{r}^e \times \Delta \boldsymbol{\sigma}_r - \Delta \boldsymbol{\sigma}_r'' \right) \\ &= \frac{\partial \mathbf{g}^e}{\partial \mathbf{r}^e} \left( \mathbf{r}^e \times \right) \Delta \boldsymbol{\sigma}_r - \frac{\partial \mathbf{g}^e}{\partial \mathbf{r}^e} \Delta \boldsymbol{\sigma}_r''. \end{aligned} \quad (33)$$

in which, the second relationship in Eq. (31) is used.

Finally, we also reformulate the differential equation as

$$\Delta \dot{\mathbf{x}}_r = \mathbf{F}_r \delta \mathbf{x}_r + \mathbf{G}_r \mathbf{w}_k. \quad (34)$$

where the error states and the Jacobian matrices are given as

$$\delta \mathbf{x}_r = \begin{bmatrix} \Delta \boldsymbol{\sigma}_r & \Delta \boldsymbol{\sigma}_r' & \Delta \boldsymbol{\sigma}_r'' & \delta \mathbf{b}_g & \delta \mathbf{b}_a \end{bmatrix}^T \quad (35)$$

$$\mathbf{F}_r = \begin{bmatrix} -\boldsymbol{\omega}_{ie}^e \times & \mathbf{0} & \mathbf{0} & \mathbf{C}_b^e & \mathbf{0} \\ \mathbf{g}^e \times - \dfrac{\partial \mathbf{g}^e}{\partial \mathbf{r}^e} \left( \mathbf{r}^e \times \right) & -\boldsymbol{\omega}_{ie}^e \times & \dfrac{\partial \mathbf{g}^e}{\partial \mathbf{r}^e} & \left( \mathbf{C}_i^e \hat{\mathbf{r}}^i \right) \times \mathbf{C}_b^e & \mathbf{C}_b^e \\ \mathbf{0} & \mathbf{I} & -\boldsymbol{\omega}_{ie}^e \times & \mathbf{r}^e \times \mathbf{C}_b^e & \mathbf{0} \\ \mathbf{0} & \mathbf{0} & \mathbf{0} & \mathbf{0} & \mathbf{0} \\ \mathbf{0} & \mathbf{0} & \mathbf{0} & \mathbf{0} & \mathbf{0} \end{bmatrix} \quad (36)$$

$$\mathbf{G}_r = \begin{bmatrix} -\mathbf{C}_b^e & \mathbf{0} & \mathbf{0} & \mathbf{0} \\ -\left( \mathbf{C}_i^e \hat{\mathbf{r}}^i \right) \times \mathbf{C}_b^e & -\mathbf{C}_b^e & \mathbf{0} & \mathbf{0} \\ -\mathbf{r}^e \times \mathbf{C}_b^e & \mathbf{0} & \mathbf{0} & \mathbf{0} \\ \mathbf{0} & \mathbf{0} & \mathbf{I} & \mathbf{0} \\ \mathbf{0} & \mathbf{0} & \mathbf{0} & \mathbf{I} \end{bmatrix} \quad (37)$$

*Remark 2:* Considering the negligible magnitude of the gravitational gradient, the first $9 \times 9$ part can be regarded as free from the state estimations. The other terms relating to $\mathbf{r}^e$ and $\hat{\mathbf{r}}^i$ cannot be omitted due to large magnitudes of the position vector ($10^6$) and the velocity vector ($10^2$). However, the attitude covariance propagation is not affected by these terms as analyzed in Appendix D. For a high-accuracy IMU, the errors introduced by inaccurate attitude estimation can be omitted as shown in Eq. (103). Therefore, the right trident quaternion error model can be regarded as free from inaccurate estimations.

In fact, the "translation" of the trident quaternion could be alternatively expressed in the body frame as

$$\bar{q}_{eb} = q_{eb} + \varepsilon_1 \frac{1}{2} q_{eb} \circ \left( \mathbf{C}_i^b \hat{\mathbf{r}}^i \right) + \varepsilon_2 \frac{1}{2} q_{eb} \circ \left( \mathbf{C}_i^b \mathbf{r}^i \right). \tag{38}$$

which is equivalent to Eq. (1) since

$$\left( \mathbf{C}_i^b \hat{\mathbf{r}}^i \right) \circ q_{eb} = q_{eb} \circ q_{eb}^* \circ \left( \mathbf{C}_i^b \hat{\mathbf{r}}^i \right) \circ q_{eb} = q_{eb} \circ \mathbf{C}_i^b \hat{\mathbf{r}}^i. \tag{39}$$

As for the robocentric representation [27], the trident quaternion can be alternatively defined as the $e$-frame relative to the $b$-frame, i.e.,

$$\bar{q}_{be} = q_{be} + \varepsilon_1 \frac{1}{2} \left( \mathbf{C}_i^b \hat{\mathbf{r}}^i \right) \circ q_{be} + \varepsilon_2 \frac{1}{2} \left( \mathbf{C}_i^b \mathbf{r}^i \right) \circ q_{be}. \tag{40}$$

The corresponding linearized left- and right-error models of robocentric formulations are directly provided in the Appendix E following similar derivation process.

## III. RELATIONS TO EXTANT STATE ERRORS

This section clarifies the relationship of newly defined trident quaternion errors w.r.t. traditional error definitions and invariant error definitions in [21], [27]. In addition, the correspondences of their covariance are also presented, which are vital in applying these methods in EKF.

### A. Relations with Traditional Error Definitions

Traditionally, the errors of the velocity and position are defined as additive errors.

$$\delta \mathbf{v}^e = \hat{\mathbf{v}}^e - \mathbf{v}^e, \tag{41}$$
$$\delta \mathbf{r}^e = \hat{\mathbf{r}}^e - \mathbf{r}^e.$$

And, the small attitude error is usually defined by small Euler angles $\Delta \boldsymbol{\sigma}_l$ or $\Delta \boldsymbol{\sigma}_r$. Subsequently, the velocity and position errors will be simultaneously defined in the left/right trident quaternion errors.

#### 1) Left errors

In contrast, the left trident quaternion errors are listed as follows (see Eqs. (9), (10) and (14))

$$\Delta q_l = \hat{q}^* \circ q \approx 1 + \frac{1}{2} \Delta \boldsymbol{\sigma}_l,$$
$$\Delta \boldsymbol{\sigma}_l' = \mathbf{C}_e^b \left( \mathbf{C}_i^e \hat{\mathbf{r}}^i - \hat{\mathbf{C}}_i^e \hat{\mathbf{r}}^i \right) = -\mathbf{C}_e^b \delta \left( \mathbf{C}_i^e \hat{\mathbf{r}}^i \right), \tag{42}$$
$$\Delta \boldsymbol{\sigma}_l'' = \mathbf{C}_e^b \left( \mathbf{r}^e - \hat{\mathbf{r}}^e \right) = -\mathbf{C}_e^b \delta \mathbf{r}^e.$$

where the left attitude error is actually the error quaternion $\Delta \sigma_l = [0, \Delta \boldsymbol{\sigma}_l]$. It can be noticed that the traditional Euler angle error $\Delta \boldsymbol{\sigma}_l$ is included in the first real part of the trident quaternion error if the error angles are small enough.

As for the covariance, we should pay attention to the relationship between $\Delta \boldsymbol{\sigma}_l'$ and $\delta \mathbf{v}^e$. Since

$$\mathbf{C}_i^e \dot{\mathbf{r}}^i = \left( \boldsymbol{\omega}_{ie}^e \times \right) \mathbf{r}^e + \mathbf{v}^e. \tag{43}$$

where $\boldsymbol{\omega}_{ie}^e = \begin{bmatrix} 0 & 0 & \omega_{ie} \end{bmatrix}^T$ is the earth rotation angular rate expressed in the earth frame. According to Eq. (42), we have

$$\begin{aligned} \mathrm{E}\left[ \Delta \boldsymbol{\sigma}_l' \Delta \boldsymbol{\sigma}_l'^T \right] &= \mathrm{E}\left[ \mathbf{C}_e^b \delta \left( \mathbf{C}_i^e \dot{\mathbf{r}}^i \right) \delta^T \left( \mathbf{C}_i^e \dot{\mathbf{r}}^i \right) \mathbf{C}_e^{bT} \right] \\ &= \mathrm{E}\left[ \mathbf{C}_e^b \left( \boldsymbol{\omega}_{ie}^e \times \delta \mathbf{r}^e + \delta \mathbf{v}^e \right) \left( \boldsymbol{\omega}_{ie}^e \times \delta \mathbf{r}^e + \delta \mathbf{v}^e \right)^T \mathbf{C}_e^{bT} \right] \\ &= \mathbf{C}_e^b \mathrm{E}\left[ \delta \mathbf{v}^e \delta \mathbf{v}^{eT} \right] \mathbf{C}_e^{bT} \\ &\quad + \mathbf{C}_e^b \left( \boldsymbol{\omega}_{ie}^e \times \right) \mathrm{E}\left[ \delta \mathbf{r}^e \delta \mathbf{r}^{eT} \right] \left( \boldsymbol{\omega}_{ie}^e \times \right)^T \mathbf{C}_e^{bT}. \end{aligned} \tag{44}$$

where 'E' denotes the expectation. Specifically, the last term can be omitted in static alignment, given the small magnitude of the earth rotation rate.

Similarly, the covariance for $\Delta \boldsymbol{\sigma}_l''$ is computed as

$$\mathrm{E}\left[ \Delta \boldsymbol{\sigma}_l'' \Delta \boldsymbol{\sigma}_l''^T \right] = \mathbf{C}_e^b \mathrm{E}\left[ \delta \mathbf{r}^e \delta \mathbf{r}^{eT} \right] \mathbf{C}_e^{bT}. \tag{45}$$

To sum up, the transformation between two covariance matrices is reformulated as

$$\bar{\mathbf{P}}_l = \mathbf{J}_l \mathbf{P} \mathbf{J}_l^T. \tag{46}$$

where $\mathbf{P}$ denotes the traditional covariance including the uncertainties of the attitude/velocity/position, and the left transformation matrix is

$$\mathbf{J}_l = \begin{bmatrix} \mathbf{I} & \mathbf{0} & \mathbf{0} \\ \mathbf{0} & \mathbf{C}_e^b & \mathbf{C}_e^b \left( \boldsymbol{\omega}_{ie}^e \times \right) \\ \mathbf{0} & \mathbf{0} & \mathbf{C}_e^b \end{bmatrix} \tag{47}$$

where $\mathbf{I}$ and $\mathbf{0}$ are 3×3 matrices.

#### 2) Right errors

In contrast, the right trident quaternion error is defined as (see Eqs. (26) and (31))

$$\Delta q_r = q \circ \hat{q}^* \approx 1 + \frac{1}{2} \Delta \sigma_r,$$
$$\Delta \boldsymbol{\sigma}_r' = \left( \mathbf{C}_i^e \hat{\mathbf{r}}^i \right) \times \Delta \boldsymbol{\sigma}_r - \delta \left( \mathbf{C}_i^e \hat{\mathbf{r}}^i \right), \tag{48}$$
$$\Delta \boldsymbol{\sigma}_r'' = \mathbf{r}^e \times \Delta \boldsymbol{\sigma}_r - \delta \mathbf{r}^e.$$

According to Eq. (43), the covariance of $\Delta \boldsymbol{\sigma}_r'$ can be given as

$$\begin{aligned} \mathrm{E}\left[ \Delta \boldsymbol{\sigma}_r' \Delta \boldsymbol{\sigma}_r'^T \right] &= \left( \mathbf{C}_i^e \hat{\mathbf{r}}^i \right) \times \mathrm{E}\left[ \Delta \boldsymbol{\sigma}_r \Delta \boldsymbol{\sigma}_r^T \right] \left[ \left( \mathbf{C}_i^e \hat{\mathbf{r}}^i \right) \times \right]^T \\ &\quad + \left( \boldsymbol{\omega}_{ie}^e \times \right) \mathrm{E}\left[ \delta \mathbf{r}^e \delta \mathbf{r}^{eT} \right] \left( \boldsymbol{\omega}_{ie}^e \times \right)^T + \mathrm{E}\left[ \delta \mathbf{v}^e \delta \mathbf{v}^{eT} \right]. \end{aligned} \tag{49}$$

Similarly, the covariance for $\Delta \boldsymbol{\sigma}_r''$ is computed as

$$\mathrm{E}\left[ \Delta \boldsymbol{\sigma}_r'' \Delta \boldsymbol{\sigma}_r''^T \right] = \left( \mathbf{r}^e \times \right) \mathrm{E}\left[ \Delta \boldsymbol{\sigma}_r \Delta \boldsymbol{\sigma}_r^T \right] \left( \mathbf{r}^e \times \right)^T + \mathrm{E}\left[ \delta \mathbf{r}^e \delta \mathbf{r}^{eT} \right], \tag{50}$$

The relationship between the traditional covariance with the newly defined covariance can be written cohesively as

$$\bar{\mathbf{P}}_r = \mathbf{J}_r \mathbf{P} \mathbf{J}_r^T. \tag{51}$$

where the transformation matrix is

$$\mathbf{J}_r = \begin{bmatrix} \mathbf{I} & \mathbf{0} & \mathbf{0} \\ \left( \mathbf{C}_i^e \hat{\mathbf{r}}^i \right) \times & \mathbf{I} & \boldsymbol{\omega}_{ie}^e \times \\ \mathbf{r}^e \times & \mathbf{0} & \mathbf{I} \end{bmatrix} \tag{52}$$

*B. Relations with Error Definitions in [38]*

Before analyzing the relationship between the current representations with those in [38], some fundamentals on Lie group and Lie algebra are introduced. According to [38], the attitude matrix $\mathbf{R}$, velocity $\mathbf{v}$ and position $\mathbf{r}$ of a rigid body can be expressed by the $SE_2(3)$ group as

$$SE_2(3) := \left\{ \chi = \begin{bmatrix} \mathbf{R} & \mathbf{v} & \mathbf{r} \\ \mathbf{0}_{1\times3} & 1 & 0 \\ \mathbf{0}_{1\times3} & 0 & 1 \end{bmatrix} \mid (\mathbf{R}, \mathbf{v}, \mathbf{r}) \in SO(3) \times \mathbb{R}^6 \right\}. \quad (53)$$

which is also called the "double direct spatial isometries" embedded in $\mathbb{R}^{6\times6}$ [21]. Besides, the group composition of two elements is defined by matrix multiplication as

$$(\mathbf{R}_1, \mathbf{v}_1, \mathbf{r}_1) \bullet (\mathbf{R}_2, \mathbf{v}_2, \mathbf{r}_2) = (\mathbf{R}_1\mathbf{R}_2, \mathbf{R}_1\mathbf{v}_2 + \mathbf{v}_1, \mathbf{R}_1\mathbf{r}_2 + \mathbf{r}_1). \quad (54)$$

Besides, the small perturbations of the Lie group is defined by the Lie algebra

$$\mathfrak{se}_2(3) := \left\{ \xi^\wedge = \begin{bmatrix} \xi^R \times & \xi^v & \xi^r \\ \mathbf{0}_{1\times3} & 1 & 0 \\ \mathbf{0}_{1\times3} & 0 & 1 \end{bmatrix}, \xi := (\xi^{R\,T}, \xi^{v\,T}, \xi^{r\,T})^T \right\}. \quad (55)$$

where the operator '$\wedge$' transforms the elements $\xi \in \mathbb{R}^9$ into the elements of Lie algebra and $\xi^R, \xi^v, \xi^r \in \mathbb{R}^3$. Moreover, the Lie algebra $\xi^\wedge$ is related with its Lie group $\chi$ in Eq. (53) by the matrix exponential

$$\chi = \exp(\xi^\wedge) = \begin{bmatrix} \exp_m(\xi^R \times) & N(\xi^R)\xi^v & N(\xi^R)\xi^r \\ \mathbf{0}_{1\times3} & 1 & 0 \\ \mathbf{0}_{1\times3} & 0 & 1 \end{bmatrix}. \quad (56)$$

In Eq. (56), the operator $\exp(\bullet)$ transforms the Lie algebra into corresponding Lie group. $\exp_m(\bullet)$ is the matrix exponential and the left Jacobian matrix satisfies

$$N(\xi^R) = \mathbf{I} + \frac{1-\cos(\|\xi^R\|)}{\|\xi^R\|^2}(\xi^R \times)^2 + \frac{\|\xi^R\| - \sin(\|\xi^R\|)}{\|\xi^R\|^3}(\xi^R \times)^3. \quad (57)$$

in which, $\|\bullet\|$ is the vector norm and the first-order approximation of $N(\xi^R)$ is

$$N(\xi^R) \approx \mathbf{I} + \xi^R \times. \quad (58)$$

Finally, with Eqs. (56)-(58) the group can be approximated up to the first order as $\xi^R, \xi^v, \xi^r$ are small.

$$\chi \approx \begin{bmatrix} \mathbf{I} + \xi^R \times & \xi^v & \xi^r \\ \mathbf{0}_{1\times3} & 1 & 0 \\ \mathbf{0}_{1\times3} & 0 & 1 \end{bmatrix}. \quad (59)$$

When the matrix Lie group was applied to design the invariant EKF on the flat Earth [21], the Earth rotation was omitted. Recently, the Earth rotation was considered in developing the exact pre-integration formulae in [38]. Rather than directly using the Earth frame velocity, the auxiliary variable $\mathbf{v}'$ was introduced to construct the "group affine" structure, defined as follows

$$\mathbf{v}' = \mathbf{v}^e + (\boldsymbol{\omega}_{ie}^e \times) \mathbf{r}^e, \quad (60)$$

Interestingly, it can be seen that Eq. (60) is exactly the velocity (cf. Eq. (43)) in the first dual part of the trident quaternion.

And, the resultant group satisfies the "group affine" property as shown in [38] (cf. Eq. (14) therein).

$$\chi = \begin{bmatrix} \mathbf{R} & \mathbf{v}' & \mathbf{r}^e \\ \mathbf{0}_{1\times3} & 1 & 0 \\ \mathbf{0}_{1\times3} & 0 & 1 \end{bmatrix}, \mathbf{R} = \mathbf{C}_b^e. \quad (61)$$

Although the authors in [38] did not use Eq. (62) to design the corresponding invariant extended Kalman filter, we will show next that the resulted nonlinear error definitions are equivalent with the currently proposed trident quaternion errors.

*1) Equivalence of Left Errors*

As in [21], [27] and [38], the left Lie group error $\eta_l \in SE_2(3)$ is defined as

$$\eta_l = \hat{\chi}^{-1}\chi = \begin{bmatrix} \hat{\mathbf{R}}^T\mathbf{R} & \hat{\mathbf{R}}^T(\mathbf{v}' - \hat{\mathbf{v}}') & \hat{\mathbf{R}}^T(\mathbf{r}^e - \hat{\mathbf{r}}^e) \\ \mathbf{0}_{1\times3} & 1 & 0 \\ \mathbf{0}_{1\times3} & 0 & 1 \end{bmatrix}. \quad (63)$$

and its corresponding Lie algebra is

$$\xi_l^\wedge = \begin{bmatrix} \xi_l^R \times & \xi_l^v & \xi_l^r \\ \mathbf{0}_{1\times3} & 1 & 0 \\ \mathbf{0}_{1\times3} & 0 & 1 \end{bmatrix} \quad (64)$$

*Proposition 1:* The left group error in Eq. (63) includes the nonlinear errors of the velocity and position, which are exactly included in the left trident quaternion error as the first and the second dual parts defined in Eq. (26).

*Proof.* Comparing Eqs. (63) and (42), the nonlinear velocity error $\hat{\mathbf{R}}^T(\mathbf{v}' - \hat{\mathbf{v}}')$ and the nonlinear position error $\hat{\mathbf{R}}^T(\mathbf{r}^e - \hat{\mathbf{r}}^e)$ are exactly equivalent to $\Delta\boldsymbol{\sigma}_l'$ and $\Delta\boldsymbol{\sigma}_l''$, respectively. In addition, the left attitude error $\xi_l^R$ is also equal to $\Delta\boldsymbol{\sigma}_l$.

*Remark 3:* In [21] and [27], the vectors $\xi_l^R, \xi_l^v, \xi_l^r$ are defined as the error states of the left IEKF. They are related with the corresponding group error to the first order as $\eta_l \approx I_d + \xi_l^\wedge$ according to Eq. (59) and (64). $I_d$ is the 5-by-5 group identity matrix. Comparing Eq. (63) with (59), the state vectors satisfy $\xi_l^R \times \approx \hat{\mathbf{R}}^T\mathbf{R} - \mathbf{I}$, $\xi_l^v \approx \hat{\mathbf{R}}^T(\mathbf{v}' - \hat{\mathbf{v}}')$, and $\xi_l^r \approx \hat{\mathbf{R}}^T(\mathbf{r}^e - \hat{\mathbf{r}}^e)$ to the first order. Similarly, the vectors $\Delta\boldsymbol{\sigma}_l$, $\Delta\boldsymbol{\sigma}_l'$ and $\Delta\boldsymbol{\sigma}_l''$ in Eq. (42) are error states of the linearized left trident quaternion kinematic model in Eq. (21). As a result, the resulted linearized kinematic models will be fully identical with the definitions of $\xi_l^R, \xi_l^v, \xi_l^r$ or $\Delta\boldsymbol{\sigma}_l, \Delta\boldsymbol{\sigma}_l', \Delta\boldsymbol{\sigma}_l''$.

*2) Equivalence of Right errors*

The right Lie group error $\eta_r \in SE_2(3)$ is denoted as

$$\eta_r = \chi\hat{\chi}^{-1} = \begin{bmatrix} \mathbf{R}\hat{\mathbf{R}}^T & \mathbf{v}' - \mathbf{R}\hat{\mathbf{R}}^T\hat{\mathbf{v}}' & \mathbf{r}^e - \mathbf{R}\hat{\mathbf{R}}^T\hat{\mathbf{r}}^e \\ \mathbf{0}_{1\times3} & 1 & 0 \\ \mathbf{0}_{1\times3} & 0 & 1 \end{bmatrix}. \quad (65)$$

and its corresponding Lie algebra is

$$\xi_r^{\wedge} = \begin{bmatrix} \boldsymbol{\xi}_r^R \times & \boldsymbol{\xi}_r^v & \boldsymbol{\xi}_r^r \\ \mathbf{0}_{1\times3} & 1 & 0 \\ \mathbf{0}_{1\times3} & 0 & 1 \end{bmatrix} \quad (66)$$

*Proposition 2:* The right group error in Eq. (65) includes the nonlinear errors of the velocity and position, which are exactly included in the right trident quaternion error as the first and the second dual parts defined in Eq. (48).

*Proof.* The right velocity error and the right position error in Eq. (65) can be approximated to the first order as

$$\mathbf{v}' - \mathbf{R}\hat{\mathbf{R}}^T\hat{\mathbf{v}}' \approx \mathbf{v}' - \left(\mathbf{I} + \boldsymbol{\xi}_r^R \times\right)\hat{\mathbf{v}}' \\ = \hat{\mathbf{v}}' \times \boldsymbol{\xi}_r^R - \delta\mathbf{v}'. \quad (67)$$

and

$$\mathbf{r}^e - \mathbf{R}\hat{\mathbf{R}}^T\hat{\mathbf{r}}^e \approx \mathbf{r}^e - \left(\mathbf{I} + \boldsymbol{\xi}_r^R \times\right)\hat{\mathbf{r}}^e \\ = \hat{\mathbf{r}}^e \times \boldsymbol{\xi}_r^R - \delta\mathbf{r}^e. \quad (68)$$

Comparing with the results in Eq. (48), the equivalence between two representations is obvious. Besides, the right attitude error $\boldsymbol{\xi}_b^R$ is also equal to $\Delta\boldsymbol{\sigma}_r$.

*Remark 4:* In [21] and [27], the vectors $\boldsymbol{\xi}_r^R, \boldsymbol{\xi}_r^v, \boldsymbol{\xi}_r^r$ are defined as the error states of the right IEKF. They are related with the corresponding group error to the first order as $\eta_r \approx I_d + \xi_r^{\wedge}$ according to Eq. (59) and (66). Comparing Eq. (65) with (59), the state vectors satisfy $\boldsymbol{\xi}_r^R \times \approx \mathbf{R}\hat{\mathbf{R}}^T - \mathbf{I}$, $\boldsymbol{\xi}_r^v \approx \mathbf{v}' - \mathbf{R}\hat{\mathbf{R}}^T\hat{\mathbf{v}}'$ and $\boldsymbol{\xi}_r^r \approx \mathbf{r}^e - \mathbf{R}\hat{\mathbf{R}}^T\hat{\mathbf{r}}^e$. Similarly, the vectors $\Delta\boldsymbol{\sigma}_r, \Delta\boldsymbol{\sigma}_r'$ and $\Delta\boldsymbol{\sigma}_r''$ in Eq. (48) are error states of the linearized right trident quaternion kinematic model in Eq. (34). As a result, the resulted linearized kinematic models will be fully identical with the definitions of $\boldsymbol{\xi}_r^R, \boldsymbol{\xi}_r^v, \boldsymbol{\xi}_r^r$ or $\Delta\boldsymbol{\sigma}_r, \Delta\boldsymbol{\sigma}_r', \Delta\boldsymbol{\sigma}_r''$.

The equivalences between trident quaternion errors and Lie group/Lie algebra errors in [38] are summarized in table I.

TABLE I
SUMMARY OF EQUIVALENCES

| Methods | Left error states | Right error states |
|---|---|---|
| $\mathfrak{se}_2(3)$ | $\boldsymbol{\xi}_l^R \times \approx \hat{\mathbf{R}}^T\mathbf{R} - \mathbf{I}$ | $\boldsymbol{\xi}_r^R \times \approx \mathbf{R}\hat{\mathbf{R}}^T - \mathbf{I}$ |
| | $\boldsymbol{\xi}_l^v \approx \hat{\mathbf{R}}^T\left(\mathbf{v}' - \hat{\mathbf{v}}'\right)$ | $\boldsymbol{\xi}_r^v \approx \mathbf{v}' - \mathbf{R}\hat{\mathbf{R}}^T\hat{\mathbf{v}}'$ |
| | $\boldsymbol{\xi}_l^r \approx \hat{\mathbf{R}}^T\left(\mathbf{r}^e - \hat{\mathbf{r}}^e\right)$ | $\boldsymbol{\xi}_r^r \approx \mathbf{r}^e - \mathbf{R}\hat{\mathbf{R}}^T\hat{\mathbf{r}}^e$ |
| $\Delta\bar{\sigma}$ | $\Delta\boldsymbol{\sigma}_l = \boldsymbol{\xi}_l^R$ | $\Delta\boldsymbol{\sigma}_r = \boldsymbol{\xi}_r^R$ |
| | $\Delta\boldsymbol{\sigma}_l' = \mathbf{C}_i^b\left(\mathbf{C}_l^e\hat{\mathbf{r}}' - \hat{\mathbf{C}}_l^e\hat{\mathbf{r}}'\right)$ | $\Delta\boldsymbol{\sigma}_r' = \left(\mathbf{C}_l^e\hat{\mathbf{r}}'\right) \times \Delta\boldsymbol{\sigma}_r - \delta\left(\mathbf{C}_l^e\hat{\mathbf{r}}'\right)$ |
| | $= \boldsymbol{\xi}_l^v$ | $= \boldsymbol{\xi}_r^v$ |
| | $\Delta\boldsymbol{\sigma}_l'' = \mathbf{C}_i^b\left(\mathbf{r}^e - \hat{\mathbf{r}}^e\right)$ | $\Delta\boldsymbol{\sigma}_r'' = \hat{\mathbf{r}}^e \times \Delta\boldsymbol{\sigma}_r - \delta\mathbf{r}^e$ |
| | $= \boldsymbol{\xi}_l^r$ | $= \boldsymbol{\xi}_r^r$ |

## IV. MEASUREMENT MODEL AND LINEARIZATION

This section first provides preliminaries about the INS/odometer integrated navigation scheme. A navigation-grade IMU is mounted on the land vehicle, and the odometer (magnetic encoder) is mounted on the non-steering wheel. As shown in Fig. 1, the center of the vehicle frame $O_m$ is situated at the middle point of the rear non-steering axle of the vehicle.

The $x_m$ axis points forward, $y_m$ axis points upward, and $z_m$ axis is along the right direction. The odometer measures the covered distance in terms of accumulated pulses, i.e., the number of pulses generated from the very start, and we assume that the measurement frame is overlapped with the vehicle frame for simplicity. The IMU frame, however, is usually misaligned with the vehicle frame by mounting angles $\varphi$, $\psi$, $\theta$. The displacement between the IMU center $O_b$ and the vehicle center $O_m$ is the lever arm $\mathbf{l}^b$, expressed in the body frame. The navigation frame is defined as north, up and east. Besides, Fig.1 also shows the relationship of the vehicle body frame w.r.t. the earth-centered earth-fixed (ECEF) frame.

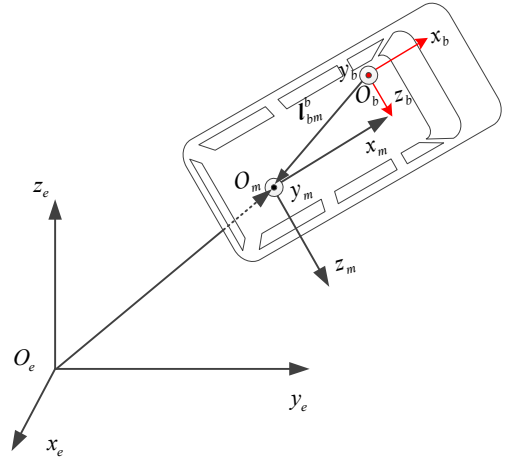

Figure 1: IMU installed on a vehicle, with definitions of the IMU body frame, the vehicle frame, the navigation frame and the ECEF frame.

As in [32] and [39], the odometer scale factor $K$ is defined as the number of pulses per meter distance. In EKF, the mounting angles, lever arms, and the scale factor are regarded as random constants. We assume that the odometer measurement frame is coincided with the vehicle frame $m$. The attitude matrix transforming the IMU body frame $b$ to the vehicle frame $m$, by the 2-3-1 rotation sequence, is given as

$$\mathbf{C}_b^m = \mathbf{M}_1(\varphi)\mathbf{M}_3(\theta)\mathbf{M}_2(\psi). \quad (69)$$

where $\mathbf{M}_i(\cdot)$ denotes the elementary rotation matrix along the $i$-th axis. The mounting angle $\varphi$ along the forward direction is unobservable [32]. Omitting this unobservable mounting angle, the parameters estimated in EKF are modeled as

$$\dot{\psi} = 0, \dot{\theta} = 0, \dot{K} = 0, \dot{\mathbf{l}}^b = \mathbf{0}, \dot{\mathbf{b}}_g = \mathbf{n}_g, \dot{\mathbf{b}}_a = \mathbf{n}_a. \quad (70)$$

where $\psi$ and $\theta$ are mounting angles along the pitch and heading directions, respectively; $K$ is the scale factor of the odometer; $\mathbf{l}^b$ is the lever arm of IMU relative to the vehicle frame.

Besides, the measurement model of the odometer pulse velocity is given as in [39].

$$y = K\mathbf{e}_1^T\mathbf{C}_b^m\left(\mathbf{C}_e^b\mathbf{v}^e + \boldsymbol{\omega}_{eb}^b \times \mathbf{l}^b\right). \quad (71)$$

where $\boldsymbol{\omega}_{eb}^b$ is the angular velocity of the vehicle body frame w.r.t. the earth frame. $\mathbf{e}_1$ is the three-dimension unit vector with the first element being 1. $\mathbf{v}^e$ is the velocity in the earth frame.

For this INS/odometer integrated navigation system, the linearized measurement models for the static alignment and the odometer-aided in-motion alignment are subsequently derived. Specifically, the static alignment relies on the zero-velocity measurement and the odometer-aided in-motion alignment depends on the land vehicle odometer pulse velocity information along with the non-holonomic constraints [39].

Based on Eqs. (22) and (35), the error states for the left/right trident quaternion extended Kalman filter (L/RQEKF) are respectively defined as

$$\Delta \mathbf{x}_l = \left[ \delta \mathbf{x}_l^T, \delta \psi, \delta \theta, \delta \mathbf{l}^{bT}, \delta K \right]^T, \tag{72}$$

and

$$\Delta \mathbf{x}_r = \left[ \delta \mathbf{x}_r^T, \delta \psi, \delta \theta, \delta \mathbf{l}^{bT}, \delta K \right]^T. \tag{73}$$

### A. Linearization of Static Alignment Measurement Model

The zero-velocity measurement model for static or quasi-static initial alignment is given as

$$\mathbf{y} = \mathbf{v}^e + \mathbf{v}_k. \tag{74}$$

where the true measurement values of $\mathbf{v}^e$ are always zero and the noise vector $\mathbf{v}_k$ accounts for the vehicle's vibrations.

And, the measurement error is defined as

$$\hat{\mathbf{y}} - \mathbf{y} = \hat{\mathbf{v}}^e - \mathbf{v}^e + \mathbf{v}_k. \tag{75}$$

#### 1) Left error model

According to the definition of the left trident quaternion error in Eq. (42) and the relationship in Eq. (43), the measurement error can be denoted as

$$\begin{aligned}
\hat{\mathbf{y}} - \mathbf{y} &= \left( \hat{\mathbf{C}}_l^e \hat{\mathbf{r}}^i - \boldsymbol{\omega}_{ie}^e \times \hat{\mathbf{r}}^e \right) - \left( \mathbf{C}_l^e \dot{\mathbf{r}}^i - \boldsymbol{\omega}_{ie}^e \times \mathbf{r}^e \right) + \mathbf{v}_k \\
&= \delta \left( \mathbf{C}_l^e \dot{\mathbf{r}}^i \right) - \boldsymbol{\omega}_{ie}^e \times \delta \mathbf{r}^e + \mathbf{v}_k \\
&= -\mathbf{C}_b^e \Delta \boldsymbol{\sigma}_l' + \boldsymbol{\omega}_{ie}^e \times \mathbf{C}_b^e \Delta \boldsymbol{\sigma}_l'' + \mathbf{v}_k.
\end{aligned} \tag{76}$$

Therefore, the measurement matrix is computed w.r.t. the left trident quaternion error as

$$\mathbf{H}_l = \begin{bmatrix} \mathbf{0}_{3\times 3} & -\mathbf{C}_b^e & \boldsymbol{\omega}_{ie}^e \times \mathbf{C}_b^e & \mathbf{0}_{3\times 12} \end{bmatrix}. \tag{77}$$

#### 2) Right error model

According to the right trident quaternion error in Eq. (48), the measurement error can be expressed as

$$\begin{aligned}
\hat{\mathbf{y}} - \mathbf{y} &= \delta \left( \mathbf{C}_l^e \dot{\mathbf{r}}^i \right) - \boldsymbol{\omega}_{ie}^e \times \delta \mathbf{r}^e + \mathbf{v}_k \\
&= \left( \mathbf{C}_l^e \dot{\mathbf{r}}^i \right) \times \Delta \boldsymbol{\sigma}_r - \Delta \boldsymbol{\sigma}_r' - \boldsymbol{\omega}_{ie}^e \times \left( \mathbf{r}^e \times \Delta \boldsymbol{\sigma}_r - \Delta \boldsymbol{\sigma}_r'' \right) + \mathbf{v}_k \\
&= \left[ \left( \mathbf{C}_l^e \dot{\mathbf{r}}^i \right) \times - \boldsymbol{\omega}_{ie}^e \times \mathbf{r}^e \times \right] \Delta \boldsymbol{\sigma}_r - \Delta \boldsymbol{\sigma}_r' + \boldsymbol{\omega}_{ie}^e \times \Delta \boldsymbol{\sigma}_r'' + \mathbf{v}_k.
\end{aligned} \tag{78}$$

Thus, the measurement matrix is computed as

$$\mathbf{H}_r = \begin{bmatrix} \left( \mathbf{C}_l^e \dot{\mathbf{r}}^i \right) \times - \boldsymbol{\omega}_{ie}^e \times \mathbf{r}^e \times & -\mathbf{I} & \boldsymbol{\omega}_{ie}^e \times & \mathbf{0}_{3\times 12} \end{bmatrix}. \tag{79}$$

**Remark 5:** In general, the linearization of the measurement matrix will affect the accuracy of covariance update in Kalman filtering, which is vital for the filtering consistency. Here, the attitude-related elements in $\mathbf{H}_l$ and $\mathbf{H}_r$ are concerned. For the left measurement matrix $\mathbf{H}_l$, its attitude-related element is $\mathbf{0}_{3\times 3}$. In contrast, the right measurement matrix $\mathbf{H}_r$ contains the estimated velocity $\hat{\mathbf{C}}_l^e \hat{\mathbf{r}}^i$. According to the derivations in Appendix F, the attitude error covariance update of the left error definition is free from inaccurate state estimates, whereas the right error definition is affected by inaccurate velocity estimate. Therefore, the left error definition is preferable for the zero-velocity measurement model.

### B. Linearization of In-motion Alignment Measurement Model

The pulse velocity measurement model of INS/odometer integrated navigation system is given as [39]

$$\mathbf{y} = diag([K \quad 1 \quad 1]) \mathbf{C}_b^m \left( \mathbf{C}_b^b \mathbf{v}^e + \boldsymbol{\omega}_{eb}^b \times \mathbf{l}^b \right). \tag{80}$$

Then the measurement error can be further approximated to the first order as

$$\begin{aligned}
\hat{\mathbf{y}} - \mathbf{y} &= \mathbf{J}_v \left( \hat{\mathbf{C}}_b^{eT} \hat{\mathbf{v}}^e - \mathbf{C}_b^{eT} \mathbf{v}^e \right) + \mathbf{J}_b \delta \mathbf{b}_g + \mathbf{J}_\psi \delta \psi \\
&\quad + \mathbf{J}_\theta \delta \theta + \mathbf{J}_l \delta \mathbf{l}^b + \mathbf{J}_K \delta K.
\end{aligned} \tag{81}$$

where coefficients matrices are specifically computed as

$$\begin{aligned}
\mathbf{J}_v &= diag([\hat{K} \quad 1 \quad 1]) \hat{\mathbf{C}}_b^m, \\
\mathbf{J}_b &= diag([\hat{K} \quad 1 \quad 1]) \hat{\mathbf{C}}_b^m \hat{\mathbf{l}}^b \times, \\
\mathbf{J}_\psi &= diag([\hat{K} \quad 1 \quad 1]) \mathbf{M}_3(\hat{\theta}) \mathbf{D}_{M_3}(\hat{\psi}) \left( \hat{\mathbf{C}}_b^{eT} \hat{\mathbf{v}}^e + \hat{\boldsymbol{\omega}}_{eb}^b \times \hat{\mathbf{l}}^b \right), \\
\mathbf{J}_\theta &= diag([\hat{K} \quad 1 \quad 1]) \mathbf{D}_{M_1}(\hat{\theta}) \mathbf{M}_2(\hat{\psi}) \left( \hat{\mathbf{C}}_b^{eT} \hat{\mathbf{v}}^e + \hat{\boldsymbol{\omega}}_{eb}^b \times \hat{\mathbf{l}}^b \right), \\
\mathbf{J}_l &= diag([\hat{K} \quad 1 \quad 1]) \hat{\mathbf{C}}_b^m \hat{\boldsymbol{\omega}}_{eb}^b \times, \\
\mathbf{J}_K &= diag([1 \quad 0 \quad 0]) \hat{\mathbf{C}}_b^m \left( \hat{\mathbf{C}}_b^{eT} \hat{\mathbf{v}}^e + \hat{\boldsymbol{\omega}}_{eb}^b \times \hat{\mathbf{l}}^b \right).
\end{aligned} \tag{82}$$

See detailed derivations presented in Appendix G.

#### 1) Left error model

In Eq. (81), the first term should be linearized w.r.t. the left trident quaternion error. According to Eq. (43), we have

$$\hat{\mathbf{C}}_b^{eT} \hat{\mathbf{v}}^e - \mathbf{C}_b^{eT} \mathbf{v}^e = -\left( \hat{\mathbf{C}}_b^{eT} \mathbf{v}^e \right) \times \Delta \boldsymbol{\sigma}_l - \Delta \boldsymbol{\sigma}_l' + \left( \hat{\mathbf{C}}_b^{eT} \boldsymbol{\omega}_{ie}^e \right) \times \Delta \boldsymbol{\sigma}_l''. \tag{83}$$

where the detailed computations are provided in Appendix H

Together with Eq. (81), the derivatives of the measurement error w.r.t. the left trident quaternion errors are

$$\begin{aligned}
\mathbf{J}_{l1} &= -\mathbf{J}_v \left( \hat{\mathbf{C}}_b^{eT} \mathbf{v}^e \right) \times, \\
\mathbf{J}_{l2} &= -\mathbf{J}_v, \\
\mathbf{J}_{l3} &= \mathbf{J}_v \left( \hat{\mathbf{C}}_b^{eT} \boldsymbol{\omega}_{ie}^e \right) \times.
\end{aligned} \tag{84}$$

Finally, the measurement matrix is formulated as

$$\mathbf{H}_l = \begin{bmatrix} \mathbf{J}_{l1}, \mathbf{J}_{l2}, \mathbf{J}_{l3}, \mathbf{J}_b, \mathbf{0}_{1\times 3}, \mathbf{J}_\psi, \mathbf{J}_\theta, \mathbf{J}_l, \mathbf{J}_K \end{bmatrix}^T. \tag{85}$$

#### 2) Right error model

The first term in Eq. (81) is linearized w.r.t. the right trident quaternion error in Eq. (48) as (see Appendix I for details)

$$\begin{aligned}
\hat{\mathbf{C}}_b^{eT} \hat{\mathbf{v}}^e - \mathbf{C}_b^{eT} \mathbf{v}^e &= -\hat{\mathbf{C}}_b^{eT} \left( \mathbf{r}^e \times \right) \left( \boldsymbol{\omega}_{ie}^e \times \right) \Delta \boldsymbol{\sigma}_r \\
&\quad - \hat{\mathbf{C}}_b^{eT} \Delta \boldsymbol{\sigma}_r' + \hat{\mathbf{C}}_b^{eT} \boldsymbol{\omega}_{ie}^e \times \Delta \boldsymbol{\sigma}_r''.
\end{aligned} \tag{86}$$

Combining with Eq. (81), the derivatives of measurement error w.r.t. the right trident quaternion errors are

$$\begin{aligned}
\mathbf{J}_{r1} &= -\mathbf{J}_v \hat{\mathbf{C}}_b^{eT} \left( \mathbf{r}^e \times \right) \left( \boldsymbol{\omega}_{ie}^e \times \right), \\
\mathbf{J}_{r2} &= -\mathbf{J}_v \hat{\mathbf{C}}_b^{eT}, \\
\mathbf{J}_{r3} &= \mathbf{J}_v \hat{\mathbf{C}}_b^{eT} \boldsymbol{\omega}_{ie}^e \times.
\end{aligned} \tag{87}$$

Finally, the measurement matrix is formulated as

$$\mathbf{H}_r = \begin{bmatrix} \mathbf{J}_{r1}, \mathbf{J}_{r2}, \mathbf{J}_{r3}, \mathbf{J}_b, \mathbf{0}_{1\times3}, \mathbf{J}_\psi, \mathbf{J}_\theta, \mathbf{J}_l, \mathbf{J}_K \end{bmatrix}^T. \quad (88)$$

*Remark 6:* As analyzed in Appendix F, the attitude-related elements in the measurement matrices are concerned. Specifically, the left measurement matrix contains the estimated attitude and velocity $\left(\hat{\mathbf{C}}_b^{eT}\hat{\mathbf{v}}^e\right)\times$, whereas the right measurement matrix only contains the estimated attitude in $\hat{\mathbf{C}}_b^{eT}\left(\mathbf{r}^e\times\right)\left(\boldsymbol{\omega}_{ie}^e\times\right)$. It is indicated that the attitude covariance update is more severely affected by the left error definition, which contains extra velocity errors. Therefore, the right error definition is better for the odometer-velocity measurement model.

### C. Realizations of L/RQEKF

For static/in-motion alignment problems, the indirect EKFs are designed based on the left/right trident quaternion error models. Specifically, the LQEKF includes the process model in Eq. (21) and the measurement models in Eq. (77) and (85). After the error states are updated, the attitude, velocity and position can be corrected by Eq. (42) as follows.

$$\hat{q}_{k|k} = \hat{q}_{k|k-1} \circ \Delta q_l,$$
$$\left(\hat{\mathbf{C}}_l^e\hat{\mathbf{r}}^l\right)_{k|k} = \mathbf{C}_{b,k|k-1}^e\Delta\boldsymbol{\sigma}_l' + \left(\hat{\mathbf{C}}_l^e\hat{\mathbf{r}}^l\right)_{k|k-1}, \quad (89)$$
$$\hat{\mathbf{r}}_{k|k}^e = \mathbf{C}_{b,k|k-1}^e\Delta\boldsymbol{\sigma}_l'' + \hat{\mathbf{r}}_{k|k-1}^e.$$

where the subscript $k|k$-1 denotes the propagated attitude/velocity/position and $k|k$ denotes the updated states.

Similarly, the RQEKF includes the process model in Eq. (34) and the measurement models in Eq. (79) and (88). The navigation states can be corrected with the updated error states by Eq. (48)

$$\hat{q}_{k|k} = \Delta q_r \circ \hat{q}_{k|k-1},$$
$$\left(\hat{\mathbf{C}}_l^e\hat{\mathbf{r}}^l\right)_{k|k} = \Delta\boldsymbol{\sigma}_r' + \left(\hat{\mathbf{C}}_l^e\hat{\mathbf{r}}^l\right)_{k|k-1} - \left(\hat{\mathbf{C}}_l^e\hat{\mathbf{r}}^l\right)_{k|k-1}\times\Delta\boldsymbol{\sigma}_r, \quad (90)$$
$$\hat{\mathbf{r}}_{k|k}^e = \Delta\boldsymbol{\sigma}_r'' + \hat{\mathbf{r}}_{k|k-1}^e - \hat{\mathbf{r}}_{k|k-1}^e\times\Delta\boldsymbol{\sigma}_r.$$

## V. SIMULATION RESULTS

Numerical simulations are conducted to show the benefits of using trident quaternion error models in the static alignment and the odometer-aided in-motion alignment. Specifically, the land vehicle is equipped with a navigation-grade IMU, which consists of a triad of gyroscopes (bias $0.005°/h$, random walks $0.001°/\text{sqrt(h)}$) and accelerometers (bias $30\mu g$, random walks $5\mu g/\text{sqrt(Hz)}$). The odometer scale factor is 59.8 p/m, i.e., roughly 59.8 pulses would be generated in 1 m distance. The IMU mounting angles are 3 deg in the yaw direction and 2 deg in the pitch direction. The lever arm is set to $\mathbf{l}^b = \begin{bmatrix} 1, 0.5, 0.8 \end{bmatrix}^T$ m. The numerical simulations and data processing are implemented in the MATLAB platform.

### A. Static Alignment

Traditionally, the analytical or optimization-based methods are used to deal with coarse alignment, hopefully better than 5 deg. Subsequently, the EKF-based fine alignment algorithm continues to improve the attitude accuracy. To imitate the practical scenarios, we assume that the uncertainties of the initial attitude error are 180 deg in three directions. This is because the unknown heading of a vehicle in the navigation frame means that the attitude in the Earth frame is nearly undetermined along three directions due to the relationship $\mathbf{C}_b^e = \mathbf{C}_n^e\mathbf{C}_b^n$. In this simulation, the navigation frame attitude error along the roll and pitch directions are presumed as 5 deg since they can be readily determined to within several degrees from the accelerometers [14], whereas the errors varying from -180 deg to 180 deg with the 5 deg gaps are added to the true heading. The initial standard deviations of attitude errors are set to 180 deg for the *e*-frame attitude along three directions to account for extremely large heading errors. Note that the body attitude is estimated in the earth-frame algorithm and then transformed to the navigation frame to compute the estimation errors. As shown in Fig. 2, the LQEKF is superior to EKF and RQEKF in terms of the convergence rate and stability. Figure 3 illustrates the root mean square errors (RMSEs) of attitude estimation errors across 73 Monte-Carlo simulations. It can be noticed that the LQEKF obviously outperforms the RQEKF and the traditional EKF and the time required to reach the 5-deg heading error is 10s and 51s for the LQEKF and RQEKF, respectively. Figure 4 plots the filtering consistency behavior under the extreme scenario of 180 deg heading error. It shows that the estimation errors of LQEKF are totally enclosed by the 3-sigma bound. As discussed in Remark 5, numerical results also indicate that the zero-velocity measurement model more prefers the left error definition, and the resultant LQEKF outperforms the EKF and RQEKF in attitude estimation accuracy and consistency. Hence, LQEKF can serve as a competitive candidate in the static alignment.

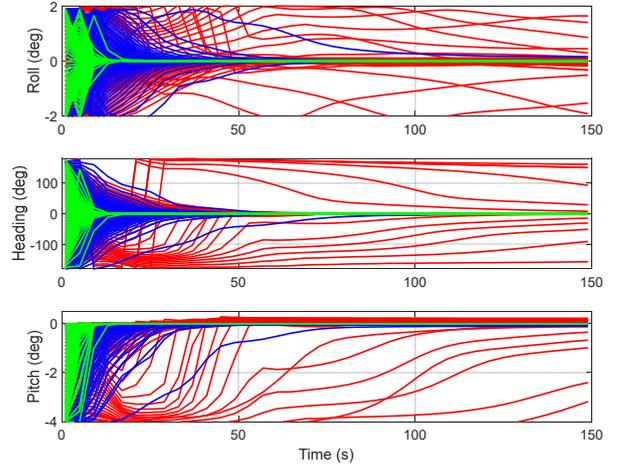

Figure 2: Monte-Carlo simulation results of attitude estimation errors. Standard EKF (red), RQEKF (blue) and LQEKF (Green).

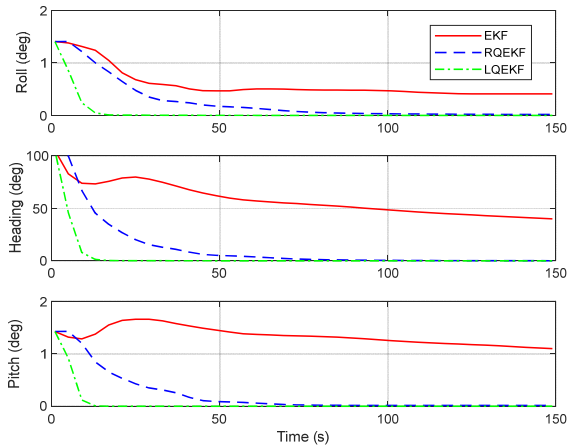

Figure 3: RMSEs of attitude estimation errors in Monte-Carlo simulations for static alignment.

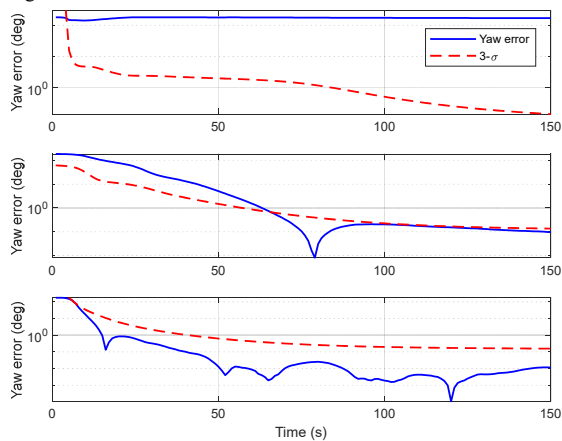

Figure 4: The consistency of three methods under 180 deg initial heading error. Dashed lines denote the 3-sigma bounds. EKF (top), RQEKF (middle), LQEKF (bottom).

### B. In-motion Alignment

The challenging in-motion alignment is performed based on the odometer pulse velocity information and the non-holonomic constraints. During the in-motion alignment process, the vehicle moves with accelerations/decelerations and turning maneuvers, which ensure the observability of state estimation. The Monte-Carlo simulations are similarly conducted when the initial heading errors range from -180 deg to 180 deg. The initial standard deviations of initial attitude errors are also set to 180 deg in three directions. Unlike the static alignment, the right linearized odometer velocity measurement model in Eq. (88) is better than (85) as analyzed in Remark 6. It is expected that the attitude convergence rate of RQEKF would be faster than LQEKF. The results of Monte-Carlo simulations are shown in Fig. 5, which indicates that all of three filtering-based methods cannot converge to satisfactory accuracy under large initial heading errors in 300 seconds. As shown in Fig. 6, the best RQEKF can only reach about 4 deg steady-state RMSE in the yaw direction with a slow convergence rate. Figure 7 shows that the RQEKF can significantly expedite the convergence under 120 deg heading error. Therefore, the L/RQEKF should be both assisted with analytical/optimization-based methods at the very start to avoid much large attitude errors.

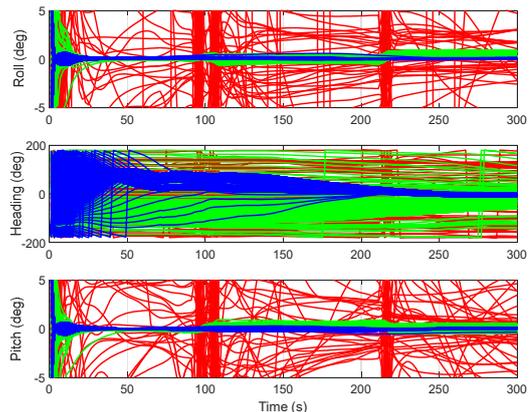

Figure 5: Monte-Carlo simulation results of attitude estimation errors for in-motion alignment. Standard EKF (red), RQEKF (blue) and LQEKF (Green).

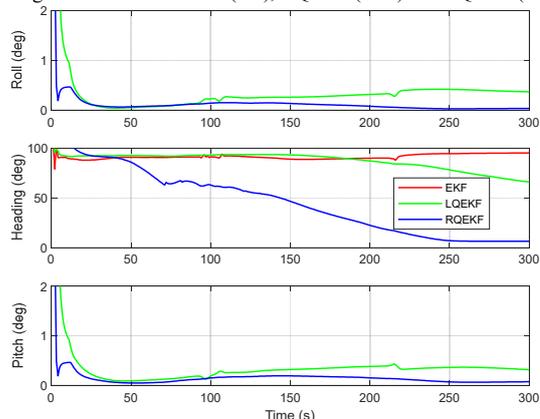

Figure 6: RMSEs of Monte-Carlo simulations of the in-motion alignment. Missed red lines for EKF means much larger RMSEs than L/RQEKF.

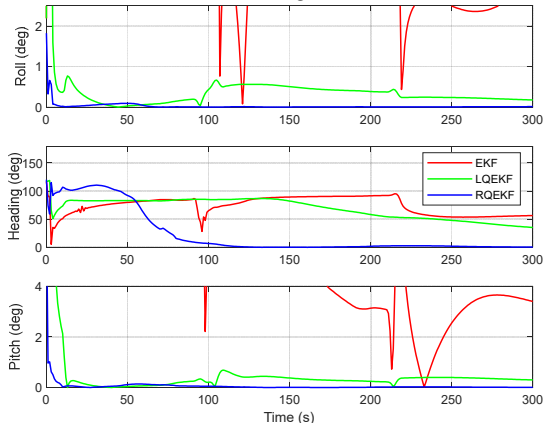

Figure 7: Attitude estimation errors under 120 deg initial heading error.

### VI. Field Test Results

Land vehicle experiments are implemented to determine the feasibility and effectiveness of applying LQEKF and RQEKF to initial alignment scenarios. The experimental system is shown in Fig. 8, in which the vehicle was equipped with a navigation-grade IMU, an odometer (wheel encoder) and a GPS receiver. Field tests are conducted at 116.4E, 39.8N. Table II lists the specifications of the system. The sampling frequency of IMU is 1kHz and the raw outputs of gyroscopes and accelerometers are downsampled to 100 Hz for the sake of quick processing. The INS/Odometer/GPS integrated navigation result starting from a static alignment is carried out and taken as the reference for subsequent assessment.

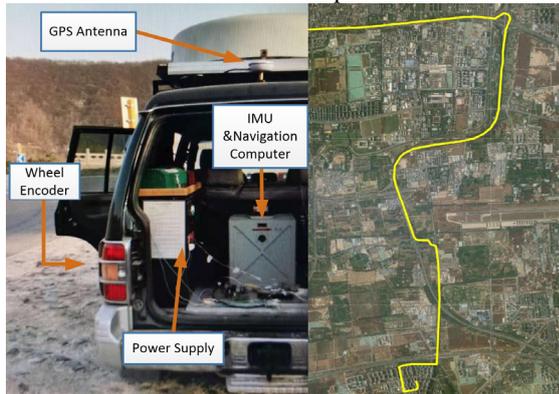

Figure 8: The experimental system and the trajectory in the field tests.

| Gyroscope | Bias | $0.01°$ / h |
| | Random Walk | $0.002°$ / sqrt(h) |
| Accelerometer | Bias | $50\mu g$ |
| | Random Walks | $10\mu g$ / sqrt(Hz) |
| Odometer | Scale factor | 53 p/m |
| GPS | Position | 5m |
| | Velocity | 0.1m/s |

### A. Static Alignment

As in the simulations, we also assume no prior information about the vehicle's initial attitude, and thus, the initial standard deviations for three directions are set to 180 deg as well. To imitate the practical scenarios, errors ranging from -180 deg to 180 deg are added to the reference heading and the 5 deg errors are added to the reference roll and pitch angles for Monte-Carlo tests. Methods are extensively verified through four INS/odometer systems of similar grade to show the stability of proposed methods. For simplicity, only the attitude estimation RMSEs of Monte-Carlo results for four static systems are provided in Figures 9-10. It can be seen that the LQEKF converges faster than the other two methods. Moreover, the final attitude estimation RMSEs at 200s are listed in Table III. It can be observed that the attitude estimation RMSEs of the LQEKF significantly outperforms the EKF, and are mostly better than the RQEKF.

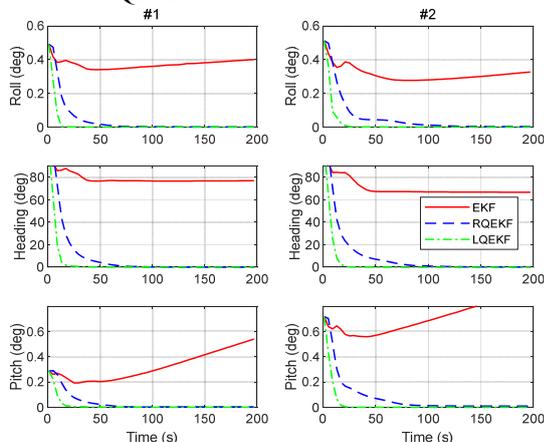

Figure 9: RMSEs of attitude errors in system #1 and #2.

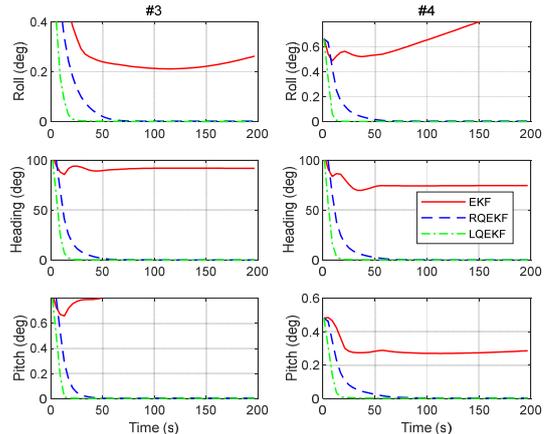

Figure 10: RMSEs of attitude errors in system #3 and #4.

| Axis | Methods | #1 (deg) | #2 (deg) | #3 (deg) | #4 (deg) |
|---|---|---|---|---|---|
| Roll | EKF | 0.401 | 0.329 | 0.266 | 0.943 |
| | RQEKF | 1.5e-3 | 2.7e-3 | 1.7e-3 | 3.1e-3 |
| | LQEKF | **5.5e-5** | **2.5e-4** | **2.5e-5** | **1.5e-4** |
| Yaw | EKF | 76.85 | 66.65 | 91.69 | 74.50 |
| | RQEKF | 0.058 | 0.168 | 0.020 | 0.026 |
| | LQEKF | **0.021** | **5.7e-3** | **0.018** | **0.019** |
| Pitch | EKF | 0.547 | 0.933 | 1.176 | 0.288 |
| | RQEKF | 3.6e-3 | 9.0e-3 | 2.3e-3 | 2.3e-3 |
| | LQEKF | **3.6e-4** | **2.5e-4** | **6.8e-5** | **1.5e-4** |

#1-4 denote the four INS/odometer systems tested.

### B. In-motion Alignment

Algorithms are tested on the trajectory shown in Fig. 8. The vehicle was driven with irregular maneuvers, including accelerations/decelerations and turnings. However, the vehicle sometimes went towards the same direction with mild accelerations/decelerations, which hinders the convergence of navigation filters. Twenty trials of in-motion alignment were randomly performed on the data of system #1 with different starting time. In each trial, the 90 deg error is added to the reference headings and the standard deviation of the initial attitude is accordingly set as 90 deg. The length of each trial is 600 seconds. Steady-state heading errors presented in Fig. 11 indicate that the performances of L/RQEKF are not stable and they even cannot converge in some cases.

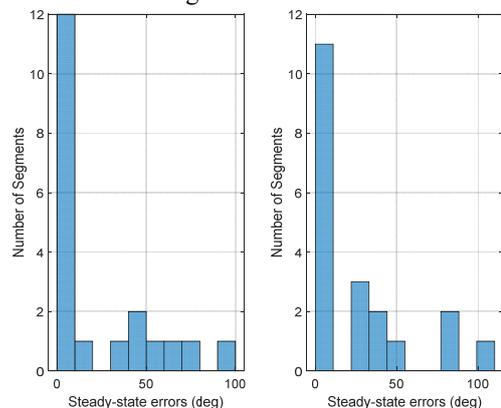

Figure 11: Statistics of steady-state heading errors: twenty trials of data in system #1 with 90 deg initial heading error: RQEKF (left), LQEKF (right).

To show the superiority of L/RQEKF over the traditional EKF, we select four representative data on which the L/RQEKF can converge under 180 deg initial heading error. The Monte-Carlo tests are conducted as those in the static alignment. The errors range from -180 deg to 180 deg are added to the true heading and the initial standard deviations are accordingly set as 180 deg in three directions. Attitude estimation RMSEs are compared in Figs. 11-12, and the corresponding steady-state value of RMSEs are listed in Table IV. These results indicate that the L/RQEKF have much better performance than the EKF, and the RQEKF converges faster and achieves higher accuracy than the LQEKF.

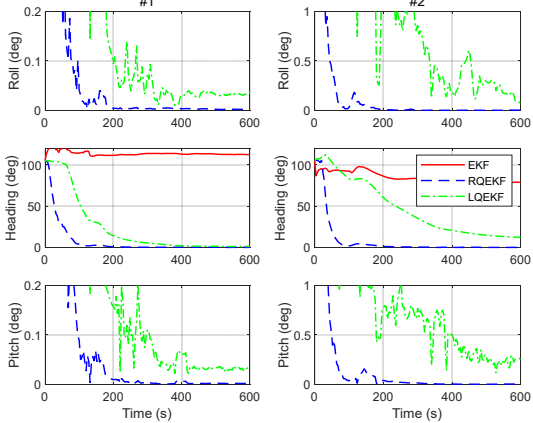

Figure 12: RMSEs of attitude errors: selected data with sufficient maneuvers from system #1 and #2.

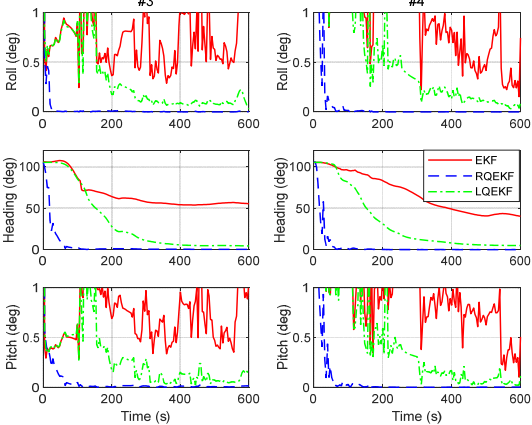

Figure 13: RMSEs of attitude errors: selected data with sufficient maneuvers from system #3 and #4.

TABLE IV
STEADY-STATE ATTITUDE RMSES OF THE IN-MOTION ALIGNMENT

| Axis | Methods | #1 (deg) | #2 (deg) | #3 (deg) | #4 (deg) |
|------|---------|----------|----------|----------|----------|
| Roll | EKF | 62.25 | 53.61 | 1.36 | 0.746 |
| | RQEKF | **2.9e-3** | **2.9e-3** | **3.6e-3** | **4.5e-4** |
| | LQEKF | 0.04 | 0.08 | 0.03 | 0.09 |
| Yaw | EKF | 112.6 | 79.19 | 55.56 | 40.22 |
| | RQEKF | **0.11** | **5.7e-2** | **0.282** | **7.4e-3** |
| | LQEKF | 1.09 | 12.20 | 4.41 | 4.99 |
| Pitch | EKF | 14.13 | 9.14 | 1.05 | 0.51 |
| | RQEKF | **1.9e-3** | **3.5e-3** | **1.4e-2** | **1.8e-3** |
| | LQEKF | 0.03 | 0.24 | 0.16 | 0.11 |

#1-4 denote the four INS/odometer systems tested.

## VII. CONCLUSIONS

Based on the trident quaternion description of the strapdown INS kinematic model, the linearized left/right-error models are accordingly derived and are found to be equivalent to those derived from the perspective of group affine. Moreover, the static and in-motion alignment algorithms for the land vehicle are studied using these error models. In applying their corresponding filtering methods, i.e., the L/RQEKF to the static alignment problem, both of them perform much better than the standard EKF. To some extent, the LQEKF might be directly used without prior attitude information in the static alignment. As for more challenging in-motion alignment scenarios, the L/RQEKF cannot ensure the convergence under extremely large initial attitude errors and their performance also depends on the practical maneuvers. It appears that the L/RQEKF have larger convergence regions than the EKF does, and the LQEKF performs relatively worse than the RQEKF in in-motion alignments. The linearized trident quaternion error models derived in this article are expected to improve the performance of other inertial-based navigation systems.

## APPENDIX A

Firstly, the errors of twists are defined as

$$\delta\hat{\omega}_{ib}^b = \hat{\omega}_{ib}^b - \bar{\omega}_{ib}^b = \delta\omega_{ib}^b + \varepsilon_1\delta f^b + \varepsilon_2\delta x_1,$$
$$\delta\hat{\omega}_{ie}^e = \hat{\omega}_{ie}^e - \bar{\omega}_{ie}^e = -\varepsilon_1\delta g^e + \varepsilon_2\delta x_2. \tag{91}$$

where the fact that $\delta\hat{\omega}_{ie}^e = 0$ is used.

Then, substitute Eq. (91) into (11)

$$\Delta\dot{\bar{\sigma}}_I = -\left(\bar{\omega}_{ib}^b + \delta\bar{\omega}_{ib}^b\right)\circ\Delta\bar{q}_I + \Delta\bar{q}_I\circ\bar{q}^*\circ\left(\bar{\omega}_{ie}^e + \delta\bar{\omega}_{ie}^e\right)\circ\bar{q}$$
$$+ \Delta\bar{q}_I\circ\left(\bar{\omega}_{ib}^b - \bar{q}^*\circ\bar{\omega}_{ie}^e\circ\bar{q}\right)$$
$$\approx -\left(\bar{\omega}_{ib}^b + \delta\bar{\omega}_{ib}^b\right)\circ\left(1 + \frac{1}{2}\Delta\bar{\sigma}_I\right) + \left(1 + \frac{1}{2}\Delta\bar{\sigma}_I\right)\circ\bar{q}^*\circ\left(\bar{\omega}_{ie}^e + \delta\bar{\omega}_{ie}^e\right)\circ\bar{q}$$
$$+ \left(1 + \frac{1}{2}\Delta\bar{\sigma}_I\right)\circ\left(\bar{\omega}_{ib}^b - \bar{q}^*\circ\bar{\omega}_{ie}^e\circ\bar{q}\right)$$
$$= -\delta\bar{\omega}_{ib}^b + \frac{1}{2}\Delta\bar{\sigma}_I\circ\bar{\omega}_{ib}^b - \frac{1}{2}\bar{\omega}_{ib}^b\circ\Delta\bar{\sigma}_I + \bar{q}^*\circ\delta\bar{\omega}_{ie}^e\circ\bar{q}$$
$$= -\delta\bar{\omega}_{ib}^b - \varepsilon_1\delta f^b - \varepsilon_2 x_1 + \left(\Delta\sigma_I + \varepsilon_1\Delta\sigma_I' + \varepsilon_2\Delta\sigma_I''\right)\times\left(\omega_{ib}^b + \varepsilon_1 f^b + \varepsilon_2 x_1\right)$$
$$+ \bar{q}^*\circ\left(-\varepsilon_1\delta g^e + \varepsilon_2\delta x_2\right)\circ\bar{q}$$
$$= -\delta\bar{\omega}_{ib}^b - \varepsilon_1\delta f^b - \varepsilon_2 x_1 - \omega_{ib}^b\times\Delta\sigma_I - \varepsilon_1 f^b\times\Delta\sigma_I - \varepsilon_1\omega_{ib}^b\times\Delta\sigma_I'$$
$$- \varepsilon_2\omega_{ib}^b\times\Delta\sigma_I'' - \varepsilon_2 x_1\times\Delta\sigma_I - \varepsilon_1 q^*\circ\delta g^e\circ q + \varepsilon_2 q^*\circ\delta x_2\circ q$$
$$= -\omega_{ib}^b\times\Delta\sigma_I - \delta\omega_{ib}^b - \varepsilon_1\left(\mathbf{C}_e^b\delta\mathbf{g}^e + \delta f^b + f^b\times\Delta\sigma_I + \omega_{ib}^b\times\Delta\sigma_I'\right)$$
$$- \varepsilon_2\left(x_1 - \mathbf{C}_e^b\delta\mathbf{x}_2 + x_1\times\Delta\sigma_I + \omega_{ib}^b\times\Delta\sigma_I''\right). \tag{92}$$

in which the following relationship has been used

$$\Delta\bar{\sigma}_I\times\bar{\omega}_{ib}^b = \frac{1}{2}\left(\Delta\bar{\sigma}_I\circ\bar{\omega}_{ib}^b - \bar{\omega}_{ib}^b\circ\Delta\bar{\sigma}_I\right). \tag{93}$$

## APPENDIX B

According to the primitive definition of the left trident quaternion error

$$\Delta\bar{q}_I = \hat{\bar{q}}^*\circ\bar{q} = \left(\hat{q}^* + \varepsilon_1\hat{q}'^* + \varepsilon_2\hat{q}''^*\right)\circ\left(q + \varepsilon_1 q' + \varepsilon_2 q''\right)$$
$$= \hat{q}^*\circ q + \varepsilon_1\left(\hat{q}'^*\circ q + \hat{q}^*\circ q\right) + \varepsilon_2\left(\hat{q}''^*\circ q + \hat{q}^*\circ q''\right) \tag{94}$$
$$\approx \Delta q_I + \varepsilon_1\frac{1}{2}\Delta\sigma_I' + \varepsilon_2\frac{1}{2}\Delta\sigma_I''.$$

We have

$$\Delta\sigma_l' = 2\Delta q_l' \circ q = 2\hat{q}'^* \circ q + 2\hat{q}^* \circ q'$$
$$= \left( \left( \hat{\mathbf{C}}_i^e \hat{\mathbf{r}}^i \right) \circ \hat{q} \right)^* \circ q + \hat{q}^* \circ \left( \mathbf{C}_i^e \mathbf{r}^i \right) \circ q$$
$$= \hat{q}^* \circ \left( \mathbf{C}_i^e \mathbf{r}^i - \hat{\mathbf{C}}_i^e \hat{\mathbf{r}}^i \right) \circ q$$
$$= -\hat{q}^* \circ q \circ q^* \circ \delta\left( \mathbf{C}_i^e \mathbf{r}^i \right) \circ q \qquad (95)$$
$$\approx -\left( 1 + \frac{1}{2}\Delta\sigma_l \right) \circ q^* \circ \delta\left( \mathbf{C}_i^e \mathbf{r}^i \right) \circ q$$
$$\approx -\mathbf{C}_e^i \delta\left( \mathbf{C}_i^e \mathbf{r}^i \right).$$

Similarly, we can prove that
$$\Delta\boldsymbol{\sigma}_l'' \approx -\mathbf{C}_e^b \delta\left( \mathbf{C}_i^e \mathbf{r}^i \right) = -\mathbf{C}_e^b \delta\mathbf{r}^e. \qquad (96)$$

## Appendix C

Similar to the derivations in Appendix A, Eq. (28) is further computed as

$$\Delta\dot{\sigma}_r = \left( \bar{q} \circ \bar{\omega}_{ib}^b \circ \bar{q}^* - \bar{\omega}_{ie}^e \right) \circ \Delta\bar{q}_r + \Delta\bar{q}_r \circ \left( \bar{\omega}_{ie}^e + \delta\bar{\omega}_{ie}^e \right)$$
$$\quad - \bar{q} \circ \left( \bar{\omega}_{ib}^b + \delta\bar{\omega}_{ib}^b \right) \circ \bar{q}^* \circ \Delta\bar{q}_r$$
$$\approx \left( \bar{q} \circ \bar{\omega}_{ib}^b \circ \bar{q}^* - \bar{\omega}_{ie}^e \right) \circ \left( 1 + \frac{1}{2}\Delta\bar{\sigma}_r \right) + \left( 1 + \frac{1}{2}\Delta\bar{\sigma}_r \right) \circ \left( \bar{\omega}_{ie}^e + \delta\bar{\omega}_{ie}^e \right)$$
$$\quad - \bar{q} \circ \left( \bar{\omega}_{ib}^b + \delta\bar{\omega}_{ib}^b \right) \circ \bar{q}^* \circ \left( 1 + \frac{1}{2}\Delta\bar{\sigma}_r \right)$$
$$\approx \frac{1}{2}\left( \Delta\bar{\sigma}_r \circ \bar{\omega}_{ie}^e - \bar{\omega}_{ie}^e \circ \Delta\bar{\sigma}_r \right) - \bar{q} \circ \delta\bar{\omega}_{ib}^b \circ \bar{q}^* + \delta\bar{\omega}_{ie}^e$$
$$= \Delta\bar{\sigma}_r \times \bar{\omega}_{ie}^e - \bar{q} \circ \delta\bar{\omega}_{ib}^b \circ \bar{q}^* + \delta\bar{\omega}_{ie}^e$$
$$= \Delta\sigma_r \times \omega_{ie}^e - q \circ \delta\omega_{ib}^b \circ q^*$$
$$\quad - \varepsilon_1 \begin{pmatrix} \omega_{ie}^e \times \Delta\sigma_r' - g^e \times \Delta\sigma_r \\ +\delta g^e + q' \circ \delta\omega_{ib}^b \circ q^* \\ +q \circ \delta\omega_{ib}^b \circ q''^* + q \circ \delta f^b \circ q^* \end{pmatrix}$$
$$\quad - \varepsilon_2 \begin{pmatrix} q'' \circ \delta\omega_{ib}^b \circ q^* + q \circ \delta\omega_{ib}^b \circ q''^* \\ +\delta x_2 + \omega_{ie}^e \times \Delta\sigma_r'' - x_2 \times \Delta\sigma_r \\ +q \circ \delta x_1 \circ q^* \end{pmatrix}$$
$$= \Delta\sigma_r \times \omega_{ie}^e - \mathbf{C}_b^e \delta\omega_{ib}^b - \varepsilon_1 \begin{pmatrix} \omega_{ie}^e \times \Delta\sigma_r' - g^e \times \Delta\sigma_r + \delta g^e \\ +\left( \mathbf{C}_i^e \mathbf{r}^i \right) \times \mathbf{C}_b^e \delta\omega_{ib}^b + \mathbf{C}_b^e \delta f^b \end{pmatrix}$$
$$\quad - \varepsilon_2 \begin{pmatrix} \mathbf{r}^e \times \mathbf{C}_b^e \delta\omega_{ib}^b + \delta x_2 - x_2 \times \Delta\sigma_r \\ +\mathbf{C}_b^e \delta\mathbf{x}_1 + \omega_{ie}^e \times \Delta\sigma_r'' \end{pmatrix}. \qquad (97)$$

In specific, we have used the following relationships in the derivations.

$$q' \circ \delta\omega_{ib}^b \circ q^* + q \circ \delta\omega_{ib}^b \circ q'^*$$
$$= q' \circ q^* \circ q \circ \delta\omega_{ib}^b \circ q^* + q \circ \delta\omega_{ib}^b \circ q^* \circ q \circ q'^*$$
$$= q' \circ q^* \circ \left( \mathbf{C}_b^e \delta\omega_{ib}^b \right) + \left( \mathbf{C}_b^e \delta\omega_{ib}^b \right) \circ q \circ q'^*$$
$$= \frac{1}{2}\left( \mathbf{C}_i^e \mathbf{r}^i \right) \circ \left( \mathbf{C}_b^e \delta\omega_{ib}^b \right) + \left( \mathbf{C}_b^e \delta\omega_{ib}^b \right) \circ q \circ \left[ \frac{1}{2}\left( \mathbf{C}_i^e \mathbf{r}^i \right) \circ q \right]^* \qquad (98)$$
$$= \frac{1}{2}\left[ \left( \mathbf{C}_i^e \mathbf{r}^i \right) \circ \left( \mathbf{C}_b^e \delta\omega_{ib}^b \right) - \left( \mathbf{C}_b^e \delta\omega_{ib}^b \right) \circ \left( \mathbf{C}_i^e \mathbf{r}^i \right) \right]$$
$$= \left( \mathbf{C}_i^e \mathbf{r}^i \right) \times \mathbf{C}_b^e \delta\omega_{ib}^b.$$

$$q'' \circ \delta\omega_{ib}^b \circ q^* + q \circ \delta\omega_{ib}^b \circ q''^* = \mathbf{r}^e \times \mathbf{C}_b^e \delta\omega_{ib}^b. \qquad (99)$$

$$\Delta\sigma_r' = 2\Delta q_r' \circ q = 2q \circ \hat{q}'^* + 2q' \circ \hat{q}^*$$
$$= q \circ \left( \left( \hat{\mathbf{C}}_i^e \hat{\mathbf{r}}^i \right) \circ \hat{q} \right)^* + \left( \mathbf{C}_i^e \mathbf{r}^i \right) \circ q \circ \hat{q}^*$$
$$\approx \left( \mathbf{C}_i^e \mathbf{r}^i \right) \circ \left( 1 + \frac{1}{2}\Delta\sigma_r \right) - \left( 1 + \frac{1}{2}\Delta\sigma_r \right) \circ \left( \mathbf{C}_i^e \mathbf{r}^i + \delta\left( \mathbf{C}_i^e \mathbf{r}^i \right) \right)$$
$$\approx \frac{1}{2}\left( \left( \mathbf{C}_i^e \mathbf{r}^i \right) \circ \Delta\sigma_r - \Delta\sigma_r \circ \mathbf{C}_i^e \mathbf{r}^i \right) - \delta\left( \mathbf{C}_i^e \mathbf{r}^i \right)$$
$$= \left( \mathbf{C}_i^e \mathbf{r}^i \right) \times \Delta\sigma_r - \delta\left( \mathbf{C}_i^e \mathbf{r}^i \right). \qquad (100)$$

## Appendix D

According to the covariance propagation process of EKF, we have

$$\mathbf{P}_{r,k|k-1} \approx \left( \mathbf{I} + \mathbf{F}_{r,k-1}\Delta t \right) \mathbf{P}_{r,k-1|k-1} \left( \mathbf{I} + \mathbf{F}_{r,k-1}\Delta t \right)^T + \mathbf{G}_{r,k-1}\mathbf{Q}_{k-1}\mathbf{G}_{r,k-1}^T \qquad (101)$$

where $\mathbf{P}_{r,k-1|k-1}$ and $\mathbf{P}_{r,k|k-1}$ denote the state covariance matrices before and after the covariance propagation; $\Delta t$ is the propagating interval. The input noise matrix is given by $\mathbf{Q}_{k-1} = \text{diag}[(\mathbf{Q}_g, \mathbf{Q}_a, \mathbf{Q}_{bg}, \mathbf{Q}_{ba})]$ , where $\mathbf{Q}_g, \mathbf{Q}_a$ denote the covariance related with random walks of gyroscopes and accelerometers, respectively; $\mathbf{Q}_{bg}, \mathbf{Q}_{ba}$ denote the covariance related with noises of gyroscopes and accelerometers biases, respectively.

For simplicity, the initial covariance matrix is assumed to be diagonal

$$\mathbf{P}_{r,k-1|k-1} = \text{diag}\left( \left[ \mathbf{P}_{\Delta\sigma_r}, \mathbf{P}_{\Delta\sigma_r'}, \mathbf{P}_{\Delta\sigma_r''}, \mathbf{P}_{bg}, \mathbf{P}_{ba} \right] \right) \qquad (102)$$

in which, $\mathbf{P}_{\Delta\sigma_r}, \mathbf{P}_{\Delta\sigma_r'}, \mathbf{P}_{\Delta\sigma_r''}, \mathbf{P}_{bg}, \mathbf{P}_{ba}$ denote the covariance w.r.t. errors of the states.

For the attitude alignment problem, we mainly focus on the accuracy of the attitude covariance propagation, i.e.,

$$\mathbf{P}_{\Delta\sigma_r} := \left( \mathbf{I} - \Delta t \boldsymbol{\omega}_{ie}^e \times \right) \mathbf{P}_{\Delta\sigma_r} \left( \mathbf{I} - \Delta t \boldsymbol{\omega}_{ie}^e \times \right)^T + \mathbf{C}_b^e \mathbf{P}_{bg} \mathbf{C}_e^b \Delta t^2 + \mathbf{C}_b^e \mathbf{Q}_g \mathbf{C}_e^b \qquad (103)$$

It can be seen that the attitude propagation only involves the estimated attitude and the introduced errors can be neglected for navigation-grade IMU.

## Appendix E

The linearized models for robocentric and previously derived world-centric representations are summarized in Table V.

TABLE V
SUMMARY OF WORLD-CENTRIC AND ROBOCENTRIC REPRESENTATIONS

| World-centric | F matrix | G matrix |
|---|---|---|
| Left-error | $\mathbf{F}_l^{wc} = \begin{bmatrix} -\omega_{ib}^b\times & 0 & 0 & I & 0 \\ -f^b\times & -\omega_{ib}^b\times & C_e^b\frac{\partial g^e}{\partial r^e}C_b^e & 0 & I \\ 0 & I & -(\omega_{ib}^b\times) & 0 & 0 \\ 0 & 0 & 0 & 0 & 0 \\ 0 & 0 & 0 & 0 & 0 \end{bmatrix}$ | $\mathbf{G}_l^{wc} = \begin{bmatrix} -I & 0 & 0 & 0 \\ 0 & -I & 0 & 0 \\ 0 & 0 & 0 & 0 \\ 0 & 0 & I & 0 \\ 0 & 0 & 0 & I \end{bmatrix}$ |
| Right-error | $\mathbf{F}_r^{wc} = \begin{bmatrix} -\omega_{ie}^e\times & 0 & 0 & C_b^e & 0 \\ g^e\times-\frac{\partial g^e}{\partial r^e}\left(r^e\times\right) & -\omega_{ie}^e\times & \frac{\partial g^e}{\partial r^e} & \left(C_b^e\dot{r}\right)\times C_b^e & C_b^e \\ 0 & I & -\omega_{ie}^e\times & r^e\times C_b^e & 0 \\ 0 & 0 & 0 & 0 & 0 \\ 0 & 0 & 0 & 0 & 0 \end{bmatrix}$ | $\mathbf{G}_r^{wc} = \begin{bmatrix} -C_b^e & 0 & 0 & 0 \\ -\left(C_b^e\dot{r}\right)\times C_b^e & -C_b^e & 0 & 0 \\ -r^e\times C_b^e & 0 & 0 & 0 \\ 0 & 0 & I & 0 \\ 0 & 0 & 0 & I \end{bmatrix}$ |

| Robocentric | F matrix | G matrix |
|---|---|---|
| Left-error | $\mathbf{F}_l^{rc} = \begin{bmatrix} -\omega_{ie}^e\times & 0 & 0 & C_b^e & 0 \\ -g^e\times-\frac{\partial g^e}{\partial r^e}\left(r^e\times\right) & -\omega_{ie}^e\times & \frac{\partial g^e}{\partial r^e} & -\left(C_b^e\dot{r}\right)\times C_b^e & C_b^e \\ 0 & I & -\omega_{ie}^e\times & -r^e\times C_b^e & 0 \\ 0 & 0 & 0 & 0 & 0 \\ 0 & 0 & 0 & 0 & 0 \end{bmatrix}$ | $\mathbf{G}_l^{rc} = \begin{bmatrix} -C_b^e & 0 & 0 & 0 \\ \left(C_b^e\dot{r}\right)\times C_b^e & -C_b^e & 0 & 0 \\ r^e\times C_b^e & 0 & 0 & 0 \\ 0 & 0 & I & 0 \\ 0 & 0 & 0 & I \end{bmatrix}$ |
| Right-error | $\mathbf{F}_r^{rc} = \begin{bmatrix} -\omega_{ib}^b\times & 0 & 0 & -I & 0 \\ f^b\times & -\omega_{ib}^b\times & C_e^b\frac{\partial g^e}{\partial r^e}C_b^e & 0 & I \\ 0 & I & -(\omega_{ib}^b\times) & 0 & 0 \\ 0 & 0 & 0 & 0 & 0 \\ 0 & 0 & 0 & 0 & 0 \end{bmatrix}$ | $\mathbf{G}_r^{rc} = \begin{bmatrix} I & 0 & 0 & 0 \\ 0 & -I & 0 & 0 \\ 0 & 0 & 0 & 0 \\ 0 & 0 & I & 0 \\ 0 & 0 & 0 & I \end{bmatrix}$ |

Superscripts *wc* denotes the world-centric versions and *rc* denotes the robocentric versions.

## Appendix F

To simplify the covariance update process of EKF, we also assume that the state covariance matrix before the update is diagonal. The covariance update process for the left measurement matrix is computed as

$$
\begin{aligned}
\mathbf{P}_{yy,l} &= \mathbf{H}_l\mathbf{P}_{k|k-1}\mathbf{H}_l^T + \mathbf{R}_k \\
&= \begin{bmatrix} \mathbf{0}_{3\times3} & -\hat{C}_b^e & \omega_{ie}^e\times\hat{C}_b^e & \mathbf{0}_{3\times12} \end{bmatrix}\text{diag}\left(\begin{bmatrix}\mathbf{P}_{\Delta\sigma_l},\mathbf{P}_{\Delta\sigma_l^\star},\cdots\end{bmatrix}\right) \\
&\quad \times \begin{bmatrix} \mathbf{0}_{3\times3} & -\hat{C}_b^e & \omega_{ie}^e\times\hat{C}_b^e & \mathbf{0}_{3\times12} \end{bmatrix}^T + \mathbf{R}_k \\
&\approx \hat{C}_b^e\mathbf{P}_{\Delta\sigma_l}\hat{C}_b^{eT} + \omega_{ie}^e\times\hat{C}_b^e\mathbf{P}_{\Delta\sigma_l^\star}\left(\omega_{ie}^e\times\hat{C}_b^e\right)^T + \mathbf{R}_k
\end{aligned}
\tag{104}
$$

$$
\begin{aligned}
\mathbf{P}_{xy,l} &= \mathbf{P}_{k|k-1}\mathbf{H}_l^T \\
&= \text{diag}\left(\begin{bmatrix}\mathbf{P}_{\Delta\sigma_l},\mathbf{P}_{\Delta\sigma_l^\star},\cdots\end{bmatrix}\right)\begin{bmatrix} \mathbf{0}_{3\times3} & -\hat{C}_b^e & \omega_{ie}^e\times\hat{C}_b^e & \mathbf{0}_{3\times12} \end{bmatrix}^T \\
&= \begin{bmatrix} \mathbf{0}_{3\times3} \\ -\mathbf{P}_{\Delta\sigma_l}\hat{C}_b^{eT} \\ \mathbf{P}_{\Delta\sigma_l^\star}\left(\omega_{ie}^e\times\hat{C}_b^e\right)^T \\ \mathbf{0}_{12\times3} \end{bmatrix}
\end{aligned}
\tag{105}
$$

$$
\begin{aligned}
\mathbf{K}_l &= \mathbf{P}_{xy,l}\mathbf{P}_{yy,l}^{-1} \\
&= \begin{bmatrix} \mathbf{0}_{3\times3} \\ -\mathbf{P}_{\Delta\sigma_l}\hat{C}_b^{eT}\mathbf{P}_{yy,l}^{-1} \\ \mathbf{P}_{\Delta\sigma_l^\star}\left(\omega_{ie}^e\times\hat{C}_b^e\right)^T\mathbf{P}_{yy,l}^{-1} \\ \mathbf{0}_{12\times3} \end{bmatrix}
\end{aligned}
\tag{106}
$$

$$\mathbf{P}_{k|k} = \left(\mathbf{I} - \mathbf{K}_l \mathbf{H}_{k,l}\right) \mathbf{P}_{k|k-1}$$
$$= \left(\mathbf{I} - \begin{bmatrix} \mathbf{0}_{3\times3} & \mathbf{0}_{3\times3} & \mathbf{0}_{3\times3} & \mathbf{0}_{3\times12} \\ \mathbf{0}_{3\times3} & * & * & \mathbf{0}_{3\times12} \\ \mathbf{0}_{3\times3} & * & * & \mathbf{0}_{3\times12} \\ \mathbf{0}_{12\times3} & \mathbf{0}_{12\times3} & \mathbf{0}_{12\times3} & \mathbf{0}_{12\times12} \end{bmatrix}\right) \mathbf{P}_{k|k-1} \quad (107)$$

where '*' denotes other trivial nonzero terms. It can be seen that the attitude covariance update is not affected by inaccurate estimations.

Following similar steps, the covariance update process for the right measurement matrix is computed as

$$\mathbf{P}_{yy,r} = \mathbf{H}_r \mathbf{P}_{k|k-1} \mathbf{H}_r^T + \mathbf{R}_k$$
$$\approx \left(\hat{\mathbf{C}}_i^e \hat{\mathbf{r}}^i\right) \times \mathbf{P}_{\Delta\sigma_r} \left(\left(\hat{\mathbf{C}}_i^e \hat{\mathbf{r}}^i\right) \times\right)^T + \mathbf{P}_{\Delta\sigma_r^e} + \mathbf{R}_k, \quad (108)$$

$$\mathbf{P}_{xy,r} = \mathbf{P}_{k|k-1} \mathbf{H}_r^T$$
$$= \begin{bmatrix} \mathbf{P}_{\Delta\sigma_r} \left(\left(\mathbf{C}_i^e \hat{\mathbf{r}}^i\right) \times -\boldsymbol{\omega}_{ie}^e \times \mathbf{r}^e \times\right)^T \\ -\mathbf{P}_{\Delta\sigma_r^e} \\ \mathbf{P}_{\Delta\sigma_r^e} \left(\boldsymbol{\omega}_{ie}^e \times\right)^T \\ \mathbf{0}_{12\times3} \end{bmatrix} \quad (109)$$

$$\mathbf{K}_r = \mathbf{P}_{xy,r} \mathbf{P}_{yy,r}^{-1}$$
$$= \begin{bmatrix} \mathbf{P}_{\Delta\sigma_r} \left(\left(\mathbf{C}_i^e \hat{\mathbf{r}}^i\right) \times -\boldsymbol{\omega}_{ie}^e \times \mathbf{r}^e \times\right)^T \mathbf{P}_{yy,r}^{-1} \\ -\mathbf{P}_v \mathbf{P}_{yy,r}^{-1} \\ \mathbf{P}_r \left(\boldsymbol{\omega}_{ie}^e \times\right)^T \mathbf{P}_{yy,r}^{-1} \\ \mathbf{0}_{12\times3} \end{bmatrix} \quad (110)$$

$$\mathbf{P}_{k|k} = \left(\mathbf{I} - \mathbf{K}_r \mathbf{H}_{k,r}\right) \mathbf{P}_{k|k-1}$$
$$\approx \left(\mathbf{I} - \begin{bmatrix} \mathbf{P}_{\Delta\sigma_r} \left(\left(\hat{\mathbf{C}}_i^e \hat{\mathbf{r}}^i\right) \times -\boldsymbol{\omega}_{ie}^e \times \hat{\mathbf{r}}^e \times\right)^T \mathbf{P}_{yy,r}^{-1} \\ \times \left(\left(\hat{\mathbf{C}}_i^e \hat{\mathbf{r}}^i\right) \times -\boldsymbol{\omega}_{ie}^e \times \hat{\mathbf{r}}^e \times\right) & * & \mathbf{0}_{3\times3} & \mathbf{0}_{3\times12} \\ * & * & * & \mathbf{0}_{3\times12} \\ * & * & * & \mathbf{0}_{3\times12} \\ \mathbf{0}_{12\times3} & \mathbf{0}_{12\times3} & \mathbf{0}_{12\times3} & \mathbf{0}_{12\times12} \end{bmatrix}\right) \mathbf{P}_{k|k-1} \quad (111)$$

where '*' denotes other trivial nonzero terms. The first $3\times3$ submatrix of $\mathbf{K}_r \mathbf{H}_r$ is mainly concerned and it can be seen that the accuracy of the attitude covariance update will be negatively affected by the estimates $\hat{\mathbf{C}}_i^e \hat{\mathbf{r}}^i$, $\hat{\mathbf{r}}^e$.

Based on the covariance update process and Eq. (111), it is indicated that the accuracy of the attitude covariance update is mainly affected by the attitude-related submatrix in the measurement matrix.

As for the odometer-velocity measurement model, the first $3\times3$ submatrices of $\mathbf{KH}$ for the left/right error definitions are respectively computed as

$$\mathbf{S}_l = \mathbf{P}_{\Delta\sigma_l} \left(\mathbf{J}_v \left(\hat{\mathbf{C}}_b^{eT} \hat{\mathbf{v}}^e\right) \times\right)^T \mathbf{P}_{yy,l}^{-1} \mathbf{J}_v \left(\hat{\mathbf{C}}_b^{eT} \hat{\mathbf{v}}^e\right) \times,$$
$$\mathbf{S}_r \approx \mathbf{P}_{\Delta\sigma_r} \left(\mathbf{J}_v \hat{\mathbf{C}}_b^{eT} \left(\mathbf{r}^e\right) \left(\boldsymbol{\omega}_{ie}^{e'}\right) \times\right)^T \mathbf{P}_{yy,r}^{-1} \mathbf{J}_v \hat{\mathbf{C}}_b^{eT} \left(\mathbf{r}^e\right) \left(\boldsymbol{\omega}_{ie}^e \times\right). \quad (112)$$

in which,

$$\mathbf{P}_{yy,l} \approx \mathbf{J}_v \left(\hat{\mathbf{C}}_b^{eT} \hat{\mathbf{v}}^e\right) \times \mathbf{P}_{\Delta\sigma_l} \left(\mathbf{J}_v \left(\hat{\mathbf{C}}_b^{eT} \hat{\mathbf{v}}^e\right) \times\right)^T + \mathbf{J}_v \mathbf{P}_{\Delta\sigma_l^i} \mathbf{J}_v^T + \mathbf{R}_k,$$
$$\mathbf{P}_{yy,r} \approx \mathbf{J}_v \hat{\mathbf{C}}_b^{eT} \left(\mathbf{r}^e\right) \left(\boldsymbol{\omega}_{ie}^e \times\right) \mathbf{P}_{\Delta\sigma_r} \left(\mathbf{J}_v \hat{\mathbf{C}}_b^{eT} \left(\mathbf{r}^e\right) \left(\boldsymbol{\omega}_{ie}^e \times\right)\right)^T$$
$$+ \mathbf{J}_v \hat{\mathbf{C}}_b^{eT} \mathbf{P}_{\Delta\sigma_r'} \left(\mathbf{J}_v \hat{\mathbf{C}}_b^{eT}\right)^T + \mathbf{R}_k. \quad (113)$$

It can be seen that $\mathbf{S}_l$ will be severely affected by the attitude error and velocity error simultaneously. In contrast, $\mathbf{S}_r$ is only affected by the attitude estimation error since the position error is negligible as compared with huge Earth radius. Hence, the right error definition introduces less errors to the attitude covariance update process.

## APPENDIX G

The odometer-velocity measurement error is approximated to the first order as

$$\hat{\mathbf{y}} - \mathbf{y} = diag([\hat{K} \quad 1 \quad 1]) \hat{\mathbf{C}}_b^m \left(\hat{\mathbf{C}}_b^{eT} \hat{\mathbf{v}}^e + \hat{\boldsymbol{\omega}}_{eb}^b \times \hat{\mathbf{l}}^b\right)$$
$$- diag([\hat{K} - \delta K \quad 1 \quad 1]) \begin{pmatrix} \hat{\mathbf{C}}_b^m - \mathbf{M}_3(\hat{\theta}) \mathbf{D}_{M_2}(\hat{\psi}) \delta\psi \\ -\mathbf{D}_{M_1}(\hat{\theta}) \mathbf{M}_2(\hat{\psi}) \delta\theta \end{pmatrix}$$
$$\times \left(\mathbf{C}_b^{eT} \mathbf{v}^e + \left(\hat{\boldsymbol{\omega}}_{eb}^b + \delta\mathbf{b}_g\right) \times \left(\hat{\mathbf{l}}^b - \delta\mathbf{l}^b\right)\right)$$
$$\approx diag([\hat{K} \quad 1 \quad 1]) \hat{\mathbf{C}}_b^m \left(\hat{\mathbf{C}}_b^{eT} \hat{\mathbf{v}}^e - \mathbf{C}_b^{eT} \mathbf{v}^e\right)$$
$$+ diag([\hat{K} \quad 1 \quad 1]) \hat{\mathbf{C}}_b^m \hat{\mathbf{l}}^b \times \delta\mathbf{b}_g$$
$$+ diag([\hat{K} \quad 1 \quad 1]) \mathbf{M}_3(\hat{\theta}) \mathbf{D}_{M_2}(\hat{\psi}) \left(\hat{\mathbf{C}}_b^{eT} \hat{\mathbf{v}}^e + \hat{\boldsymbol{\omega}}_{eb}^b \times \hat{\mathbf{l}}^b\right) \delta\psi$$
$$+ diag([\hat{K} \quad 1 \quad 1]) \mathbf{D}_{M_1}(\hat{\theta}) \mathbf{M}_2(\hat{\psi}) \left(\hat{\mathbf{C}}_b^{eT} \hat{\mathbf{v}}^e + \hat{\boldsymbol{\omega}}_{eb}^b \times \hat{\mathbf{l}}^b\right) \delta\theta$$
$$+ diag([\hat{K} \quad 1 \quad 1]) \hat{\mathbf{C}}_b^m \hat{\boldsymbol{\omega}}_{eb}^b \times \delta\mathbf{l}^b$$
$$+ diag([1 \quad 0 \quad 0]) \hat{\mathbf{C}}_b^m \left(\hat{\mathbf{C}}_b^{eT} \hat{\mathbf{v}}^e + \hat{\boldsymbol{\omega}}_{eb}^b \times \hat{\mathbf{l}}^b\right) \delta K$$
$$\triangleq \mathbf{J}_v \left(\hat{\mathbf{C}}_b^{eT} \hat{\mathbf{v}}^e - \mathbf{C}_b^{eT} \mathbf{v}^e\right) + \mathbf{J}_b \delta\mathbf{b}_g + \mathbf{J}_\psi \delta\psi + \mathbf{J}_\theta \delta\theta + \mathbf{J}_l \delta\mathbf{l}^b + \mathbf{J}_K \delta K. \quad (114)$$

where the approximation $\delta\boldsymbol{\omega}_{eb}^b = \delta\boldsymbol{\omega}_{ib}^b - \delta\mathbf{C}_e^b\boldsymbol{\omega}_{ie}^e \approx -\delta\mathbf{b}_g$ is used. And, the IMU-vehicle misalignment matrix and the derivatives of the elementary rotation matrix are given as

$$\mathbf{C}_b^m = \mathbf{M}_3(\theta) \mathbf{M}_2(\psi), \quad (115)$$

$$\mathbf{D}_{M_2}(\psi) = \begin{bmatrix} -\sin\psi & 0 & -\cos\psi \\ 0 & 0 & 0 \\ \cos\psi & 0 & -\sin\psi \end{bmatrix}, \mathbf{D}_{M_1}(\theta) = \begin{bmatrix} -\sin\theta & \cos\theta & 0 \\ -\cos\theta & -\sin\theta & 0 \\ 0 & 0 & 0 \end{bmatrix}. \quad (116)$$

## APPENDIX H

Note that $\hat{\mathbf{C}}_b^e = \mathbf{C}_b^e$ in this place, and the derivations of Eq. (83) are given as

$$
\begin{aligned}
\hat{\mathbf{C}}_b^{eT} \hat{\mathbf{v}}^e - \mathbf{C}_b^{eT} \mathbf{v}^e &= \mathbf{C}_b^{eT} \hat{\mathbf{v}}^e - \left( \mathbf{C}_b^e \mathbf{C}_b^{\hat{b}} \right)^T \mathbf{v}^e \\
&= \mathbf{C}_b^{eT} \hat{\mathbf{v}}^e - \mathbf{C}_b^{\hat{b}T} \mathbf{C}_b^{eT} \mathbf{v}^e.
\end{aligned}
\tag{117}
$$

According to Eq. (42), Eq. (83) could be further computed as

$$
\begin{aligned}
\hat{\mathbf{C}}_b^{eT} \hat{\mathbf{v}}^e - \mathbf{C}_b^{eT} \mathbf{v}^e &= \hat{\mathbf{C}}_b^{eT} \hat{\mathbf{v}}^e - \left( \mathbf{I} + \Delta \boldsymbol{\sigma}_l \times \right)^T \hat{\mathbf{C}}_b^{eT} \mathbf{v}^e \\
&= \hat{\mathbf{C}}_b^{eT} \delta \mathbf{v}^e - \left( \hat{\mathbf{C}}_b^{eT} \mathbf{v}^e \right) \times \Delta \boldsymbol{\sigma}_l \\
&= \hat{\mathbf{C}}_b^{eT} \left( \delta \left( \mathbf{C}_i^e \dot{\mathbf{r}}^i \right) - \boldsymbol{\omega}_{ie}^e \times \delta \mathbf{r}^e \right) - \left( \hat{\mathbf{C}}_b^{eT} \mathbf{v}^e \right) \times \Delta \boldsymbol{\sigma}_l \\
&= -\Delta \boldsymbol{\sigma}_l' - \left( \hat{\mathbf{C}}_b^{eT} \boldsymbol{\omega}_{ie}^e \right) \times \hat{\mathbf{C}}_b^{eT} \delta \mathbf{r}^e - \left( \hat{\mathbf{C}}_b^{eT} \mathbf{v}^e \right) \times \Delta \boldsymbol{\sigma}_l \\
&= -\left( \hat{\mathbf{C}}_b^{eT} \mathbf{v}^e \right) \times \Delta \boldsymbol{\sigma}_l - \Delta \boldsymbol{\sigma}_l' + \left( \hat{\mathbf{C}}_b^{eT} \boldsymbol{\omega}_{ie}^e \right) \times \Delta \boldsymbol{\sigma}_l''.
\end{aligned}
\tag{118}
$$

## APPENDIX I

Note that $\hat{\mathbf{C}}_b^e = \mathbf{C}_b^e$ in this place, and derivations of Eq. (86) are provided based on Eqs. (43) and (48),

$$
\begin{aligned}
& \hat{\mathbf{C}}_b^{eT} \hat{\mathbf{v}}^e - \mathbf{C}_b^{eT} \mathbf{v}^e \\
&= \hat{\mathbf{C}}_b^{eT} \left( \hat{\mathbf{v}}^e - \mathbf{C}_b^{\hat{e}} \mathbf{C}_b^{eT} \mathbf{v}^e \right) \\
&\approx \hat{\mathbf{C}}_b^{eT} \left( \mathbf{C}_i^e \dot{\mathbf{r}}^i - \boldsymbol{\omega}_{ie}^e \times \mathbf{r}^e - \left( \mathbf{I} + \Delta \boldsymbol{\sigma}_r \times \right)^T \left( \mathbf{C}_i^e \dot{\mathbf{r}}^i - \boldsymbol{\omega}_{ie}^e \times \mathbf{r}^e \right) \right) \\
&= \hat{\mathbf{C}}_b^{eT} \left( \delta \left( \mathbf{C}_i^e \dot{\mathbf{r}}^i \right) - \left( \mathbf{C}_i^e \dot{\mathbf{r}}^i \right) \times \Delta \boldsymbol{\sigma}_r - \boldsymbol{\omega}_{ie}^e \times \delta \mathbf{r}^e + \left( \boldsymbol{\omega}_{ie}^e \times \mathbf{r}^e \right) \times \Delta \boldsymbol{\sigma}_r \right) \\
&= \hat{\mathbf{C}}_b^{eT} \left[ \left( \boldsymbol{\omega}_{ie}^e \times \mathbf{r}^e \right) \times - \left( \boldsymbol{\omega}_{ie}^e \times \right) \left( \mathbf{r}^e \times \right) \right] \Delta \boldsymbol{\sigma}_r - \hat{\mathbf{C}}_b^{eT} \Delta \boldsymbol{\sigma}_r' + \hat{\mathbf{C}}_b^{eT} \boldsymbol{\omega}_{ie}^e \times \Delta \boldsymbol{\sigma}_r'' \\
&= -\hat{\mathbf{C}}_b^{eT} \left( \mathbf{r}^e \times \right) \left( \boldsymbol{\omega}_{ie}^e \times \right) \Delta \boldsymbol{\sigma}_r - \hat{\mathbf{C}}_b^{eT} \Delta \boldsymbol{\sigma}_r' + \hat{\mathbf{C}}_b^{eT} \boldsymbol{\omega}_{ie}^e \times \Delta \boldsymbol{\sigma}_r''.
\end{aligned}
\tag{119}
$$

where the following relationship is used

$$
\left( \mathbf{a} \times \mathbf{b} \right) \times = \left( \mathbf{a} \times \right) \left( \mathbf{b} \times \right) - \left( \mathbf{b} \times \right) \left( \mathbf{a} \times \right)
\tag{120}
$$